%% file: main.tex
\documentclass{article} % For LaTeX2e
\usepackage{iclr2027_conference,times}

\input{math_commands.tex}

\usepackage{hyperref}
\usepackage{url}

\usepackage{booktabs}       % professional-quality tables
\usepackage{amsfonts}       % blackboard math symbols
\usepackage{nicefrac}       % compact symbols for 1/2, etc.
\usepackage{microtype}      % microtypography
\usepackage{xcolor}         % colors
\usepackage{footnotebackref}
\usepackage{amsmath,amsfonts}
\usepackage{graphicx}
\usepackage{bm}
\usepackage{algorithm, algpseudocode}
\usepackage{wrapfig}
\usepackage{multirow}

\newcommand{\bpval}[1]{\textcolor{gray}{$#1$}}
\newcommand{\bplead}[1]{\textcolor{gray}{$\mathit{#1}$}}

\usepackage{array}
\usepackage{amsthm}
\usepackage{mdframed}
\theoremstyle{definition}   % upright body text, so the theorem's table stays legible
\newtheorem{theorem}{Theorem}
\newenvironment{boxedtheorem}
  {\begin{mdframed}[linewidth=0.8pt, innertopmargin=8pt, innerbottommargin=8pt,
                    innerleftmargin=10pt, innerrightmargin=10pt, skipabove=8pt, skipbelow=8pt]%
   \begin{theorem}}
  {\end{theorem}\end{mdframed}}

\title{Correlations Are Ruining Your \\Gradient Descent}

\author{Joshua Offergeld \& Nasir Ahmad \\
Donders Institute for Brain, Cognition, and Behaviour\\
Radboud University, Nijmegen, the Netherlands\\
\texttt{joshua.offergeld@ru.nl, nasir.ahmad@donders.ru.nl} \\
}

\iclrfinalcopy % Uncomment for camera-ready version, but NOT for submission.
\begin{document}

\maketitle

\input{sections/abstract_arxiv}

\input{sections/intro}

\input{sections/methods}

\input{sections/results}

\input{sections/discussion}

\bibliography{iclr2027_conference}
\bibliographystyle{iclr2027_conference}

\appendix

\input{sections/appendices}

\end{document}

%% file: math_commands.tex
\usepackage{amsmath,amsfonts,bm}

\def\eqref#1{equation~\ref{#1}}
\def\1{\bm{1}}

\DeclareMathAlphabet{\mathsfit}{\encodingdefault}{\sfdefault}{m}{sl}
\SetMathAlphabet{\mathsfit}{bold}{\encodingdefault}{\sfdefault}{bx}{n}

%% file: sections/abstract_arxiv.tex
\begin{abstract}
Natural gradient descent, and the preconditioned optimizers which have followed from it, correct gradient descent by estimating curvature and applying its inverse to the gradient.
For any single layer of a deep neural network, this correction can be shown to approximately separate into two factors: one which accounts for correlations in the gradient signal, and one which accounts for correlations in the data arriving at that layer.
Herein we show that this second factor can be expressed as a change of basis rather than of curvature and, considered thus, it can be removed from the optimizer entirely and placed within the network itself.
Specifically, we show that, once forward weights are corrected, right-sided natural gradient descent can be implemented in a form which has three exactly equivalent forms: a gradient preconditioner, a decorrelating linear transform in the forward pass, or a set of recurrent dynamics in which all updates are Hebbian and rely only upon locally available signals.
The (recurrent) decorrelation method we discover aligns closely with proposed methods for inhibitory learning and homeostatic plasticity in the brain.
Notably, our transform requires no matrix inverse, no matrix decomposition, and no information beyond that which is available at an individual synapse.
It can therefore be applied in distributed and physically constrained systems, including neuromorphic hardware and nervous systems, where existing preconditioners cannot.
We demonstrate its effectiveness empirically in convolutional and residual network models while significantly improving a range of approximate biologically plausible methods for credit assignment. Code is available at \href{https://github.com/EfficientNeuralSystems/correlations-ruin-gradient-descent}{https://github.com/EfficientNeuralSystems/correlations-ruin-gradient-descent}.
\end{abstract}

%% file: sections/intro.tex
\section{Introduction}
The method of gradient descent is as popular as it is intuitive.
Nonetheless, a gradient is only the `steepest direction' with respect to some measure of distance in parameter space, and gradient descent makes a particular, and rarely examined, choice of that measure.
Much of the recent progress in optimization has come from revisiting this choice, and methods which precondition the gradient \citep{Martens2015-tx, Gupta2018-co, Vyas2024-bh, jordan2024muon} have thereby moved from the edges of the field to its centre.

Natural gradient descent \citep{Amari1998-wh} is the oldest of these perspectives, measuring distance by a Riemannian metric rather than a Euclidean one.
Applied to a single layer of a deep neural network, its correction separates, approximately, into two factors: one (left) built from correlations in the gradient signal arriving at that layer, and one (right) built from correlations in the data entering it \citep{Heskes2000-om, Martens2015-tx}.
% The left factor, which undoes the skew of the loss landscape, has received almost all of the attention.
While the left factor undoes the skew of the loss landscape, the right factor is the subject of this work, and our starting observation is that it is not really a statement about curvature at all.
Correlations in the data arriving at a layer mean only that the layer's parameters live in a non-orthonormal basis, and the right factor is precisely the transformation which returns them to an orthonormal one.
This problem of basis, unlike a problem of curvature, doesn't need to be solved by an optimizer but can be tracked and solved within a network outside of training.

\subsection{Gradient descent, natural gradients, and the choice of metric}
\label{sec:bg_metric}
\label{sec:bg_dnn}

For a dataset $\mathcal{D}$ of input and output pairs $(\bm{x},\bm{y})$, we wish to find parameters $\bm{\theta}$ of a function $\bm{f}_{\bm{\theta}}$ which minimise a loss $\mathcal{L}(\bm{\theta}) = \frac{1}{|\mathcal{D}|}\sum_{(\bm{x},\bm{y})\in\mathcal{D}} \ell(\bm{f}_{\bm{\theta}}(\bm{x}), \bm{y})$, with a sample-wise loss $\ell$.
Gradient descent proposes the update $\delta\bm{\theta}^{\text{GD}} = -\eta\nabla\mathcal{L}(\bm{\theta})$, and though this is usually described as a `small step' in the direction of steepest descent, it is equivalently the minimum of a linearised loss which is penalised by the size of the parameter change,
\begin{equation*}
\delta {\bm{\theta}}^{\text{GD}} = \underset{\delta \bm{\theta}}{\operatorname{argmin}} \, \mathcal{L}(\bm{\theta}) + \nabla \mathcal{L}(\bm{\theta})^\top \delta \bm{\theta} + \frac{1}{2 \eta}{\delta \bm{\theta}}^\top \delta \bm{\theta}.
\end{equation*}
The final term is the Euclidean norm of the parameter change, and it is here that gradient descent makes its choice of metric.

\citet{Amari1998-wh} proposed that this choice be replaced by a general Riemannian metric, ${\delta \bm{\theta}}^\top \bm{G}(\bm{\theta}) \delta \bm{\theta}$, whereupon minimisation yields the natural gradient descent update
\begin{equation}
{\delta \bm{\theta}}^{\text{NGD}} = - \eta \bm{G}(\bm{\theta})^{-1} \nabla \mathcal{L}(\bm{\theta}).
\label{eq:bg_ngd}
\end{equation}
Standard natural gradient descent takes $\bm{G}$ to be the Fisher information matrix, which assumes the problem to be probabilistic.
We instead follow \citet{Heskes2000-om}, who use a deterministic perspective to derive the Fisher, though one may alternatively consider the change induced in the loss, and thereby obtain, $\bm{G} = \left \langle \nabla_{\bm{\theta}} \ell \; \nabla_{\bm{\theta}} \ell^{\top} \right \rangle_{\bm{x},\bm{y}}$, where $\langle \cdot \rangle_a$ denotes the expectation over the variable $a$. The derivation is provided in Appendix~\ref{app:lossdist}.

Note that this metric, being an outer product of gradients, is a well established quantity commonly referred to as the \textit{empirical} Fisher information matrix.
% It is well established that this is not in general a reliable proxy for the curvature of a loss landscape \citep{kunstner2019limitations, Martens2020-oo}, and we make no claim that it is.
As will be seen below, under a Kronecker factorization the factor built from a layer's input data is identical whether one begins from the empirical Fisher, the Fisher, or the generalized Gauss-Newton matrix\footnote{For a discussion of the (empirical) Fisher, and the connection to the generalized Gauss-Newton matrix, we refer to \citep{kunstner2019limitations}.}; these differ only in the factor built from the gradient signal.
Since the method we develop acts upon the input factor alone, and discards the other entirely, the known deficiencies of the empirical Fisher can be ignored.

For a feed-forward network $\bm{x}_i = \phi(\bm{h}_i) = \phi(\bm{W}_i \bm{x}_{i-1})$ over layers $i \in [1 \dots L]$, with $\phi$ a fixed continuous non-linearity, $\bm x_0$ the input data, $\bm x_L$ the model output and biases omitted throughout, it is standard to assume that the natural gradient update may be computed for each layer independently, an approximation which works well in practice \citep{desjardins2015natural} and is fully justified for linear networks \citep{Bernacchia2018-zj}.
Under this assumption, and that the gradient signal and the input data may be separately averaged, the metric for a single layer factorises as a Kronecker product of two terms and the update becomes
\begin{equation}\label{eq:bg_NG_DNN}
    {\delta \bm{\theta}}^{\text{NG}}_{\bm{W}_i}
    = - {\eta} \text{Vec} \left( \left \langle \frac{\partial \ell}{\partial \bm{h}_i} \frac{\partial \ell}{\partial \bm{h}_i}^\top \right \rangle^{-1}_{\bm{y}} \left \langle \frac{\partial \ell}{\partial \bm{h}_i} \bm{x}_{i-1}^\top \right \rangle_{\bm{y}, \bm{x}_{i-1}} \left \langle \bm{x}_{i-1} \bm{x}_{i-1}^\top \right \rangle^{-1}_{\bm{x}_{i-1}} \right ), \\
\end{equation}
where $\text{Vec}(\cdot)$ flattens a matrix into a vector.
This result has been arrived at independently by \citet{Heskes2000-om}, \citet{Martens2015-tx}, \citet{desjardins2015natural} and \citet{Bernacchia2018-zj}; the full derivation is reproduced in Appendix~\ref{app:dnn_natgrads}.

The structure of Equation~\ref{eq:bg_NG_DNN} is the starting point for everything which follows.
In the middle sits the ordinary gradient, as computed by backpropagation.
From the left it is multiplied by the inverse correlation of the gradient vectors, which undoes skew and scale in the loss landscape and is the term with which natural gradients are usually associated.
From the right it is multiplied by the inverse correlation of the data which acts as input to that layer, and it is this right factor alone which this work targets.
Note that we refer to these outer product terms as correlations throughout, despite the fact that a true correlation would require a centering and a normalization of the data, and we do so purely for ease of discussion.

\subsection{Preconditioned optimizers and what they require}
\label{sec:bg_precond}

Equation~\ref{eq:bg_NG_DNN} is rarely applied directly, as both factors must be estimated and inverted.
K-FAC \citep{Martens2015-tx} estimates the two Kronecker factors and inverts them, refreshing these inverses only every several hundred steps in order that the cost be amortised.
Shampoo \citep{Gupta2018-co} instead maintains matrix roots of preconditioners built from the gradients themselves, and SOAP \citep{Vyas2024-bh} runs an adaptive method within the eigenbasis which such a preconditioner provides.
Muon \citep{jordan2024muon} dispenses with curvature estimation altogether and orthogonalizes the gradient matrix directly, which amounts to steepest descent under a spectral rather than a Euclidean norm.
Recently proposed one-sided variants of these methods~\citep{an2026asgo} precondition on a single side only, and thereby approach the term which we study here from the direction of the optimizer.

Two properties are common to all of these methods.
First, each depends upon statistics of the gradient, which are available only once a backward pass has been computed.
Second, each requires a matrix inverse, decomposition, or iterative orthogonalization, which gathers the statistics of an entire layer into one place and processes them by expensive machinery which sits outside of the network.
The right factor of Equation~\ref{eq:bg_NG_DNN} is unusual in requiring neither: it is built from a layer's inputs, and so is available during the forward pass alone.

\subsection{Decorrelation and whitening of layer inputs}
\label{sec:bg_decor}

That the right factor may be addressed by acting upon a network's activations rather than upon its gradients is not a new idea, and two families of method exist for doing so.
The first measures the correlation matrix of the data at a layer and inverts it by matrix decomposition, whether at pre-defined checkpoints \citep{desjardins2015natural, Luo2017-ku}, at the level of individual mini-batches \citep{Bernacchia2018-zj}, or as an extension of batch normalization applied to networks of considerable depth \citep{Huang2018-zy, Huang2019-ms}.
Of these, \citet{desjardins2015natural} alone recognised that re-estimating such a transform alters the function which the network computes, and re-parameterised the downstream weights in order to preserve it.
The second family estimates a decorrelating matrix iteratively, with no inversion or decomposition at any point \citep{silva1991distributed, silva1991speeding, Ahmad2022-hv, Dalm2024-ly, Sussillo2009-ws}.
These methods are cheap and they are online, but none of them corrects the downstream weights, and none has been linked to natural gradients.
A third and parallel literature, in computational neuroscience, has derived recurrent and local dynamics which decorrelate the activity of a population of neurons; we return to it in Section~\ref{sec:discussion}, as our recurrent formulation turns out to resemble it closely.

Approximate methods for credit assignment, which replace backpropagation with signals that a distributed system might plausibly transport, are well known to fail upon tasks of even moderate difficulty \citep{bartunov2018assessing}, and it has been suggested that their choice of metric is partly to blame \citep{surace2020choice}.
Notably, such methods, being approximate methods, by definition contain errors in the gradient signal, however their forward pass computations are not approximate.
The factorization above therefore predicts that the right factor is exactly that part of the natural gradient which remains repairable within networks and without resorting to gradient computations.

\subsection*{Contributions}

In this work we uncover each of these aspects of right-sided natural gradient descent and contribute the following:
\begin{enumerate}
    \item We show an implementation of right-sided natural gradient descent which, when forward weights are corrected, has three exactly equivalent forms: a gradient preconditioner, a decorrelating linear transform in the forward pass, or a set of recurrent dynamics with purely Hebbian updates.
    \item We show that this transform can be learned by a rank-one update, requiring no matrix inverse, no matrix decomposition, and no information beyond that which is presently available at an individual synapse. It therefore could be applied in distributed and physically constrained systems in which existing preconditioners cannot be implemented.
    \item We derive the correcting update to the forward weights and show it to be the element missing from existing methods of online decorrelation \citep{silva1991distributed, Ahmad2022-hv, Dalm2024-ly, Huang2018-zy}. Without it, every decorrelation step displaces the network upon its loss landscape by an amount which is neither measured nor controlled.
    \item We demonstrate empirically that our method matches or outperforms Shampoo and Muon when paired with backpropagation, and that it significantly improves a range of approximate methods for credit assignment.
\end{enumerate}

%% file: sections/methods.tex
\section{Methods}\label{sec:decor}

\subsection{Data correlations place a layer's parameters in a non-orthonormal basis}
\label{sec:nonorthonormal}

% It is ultimately rather trivial to show why data correlations ruin gradient descent.
To show why data correlations ruin gradient descent, consider a set of data with some correlation structure, $\langle \bm{x} \bm{x}^\top \rangle = \bm{\Sigma}$.
Supposing this data to be well conditioned, one may equally write it as a whitened dataset multiplied by the matrix square root of its correlation matrix, $\bm{x} = \bm{\Sigma}^{1/2}\bar{\bm{x}}$, where $\langle \bar{\bm{x}} \bar{\bm{x}}^\top \rangle = \bm{I}$.
The output of a linear model of these inputs may then be written $\hat{\bm{z}} = \bm{W}\bm{x} = (\bm{W}\bm{\Sigma}^{1/2})\bar{\bm{x}}$, in which the correlation matrix appears as a matrix by which our parameters are brought into a non-orthonormal relationship.\footnote{Note that we specifically use the term \textit{orthonormal} rather than \textit{orthogonal}. Because our derivation is based on a loss-distance metric, both the cross-terms (correlations) and the diagonal terms (variances) of the data define the parameter impact. Thus, properly preconditioning the gradients requires both orthogonalizing the features and normalizing their variances to exactly restore an orthonormal basis.}
Thus, if we compute gradient descent with respect to $\bm{W}$ without accounting for the correlation inducing matrix $\bm{\Sigma}^{1/2}$, we update parameters which no longer stand in an orthonormal relationship to one another.

\begin{figure}[!ht]
    \centering
    \includegraphics[width=0.7\linewidth]{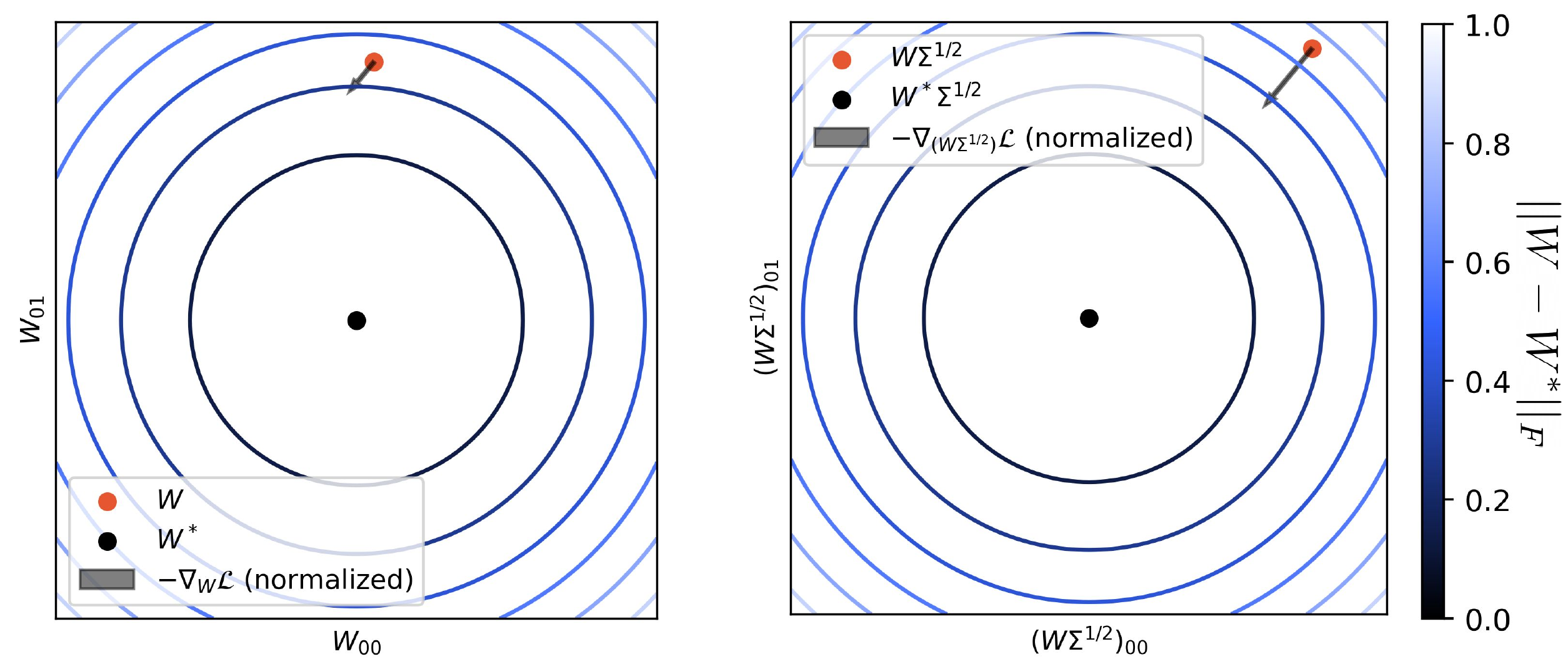}
    \caption{Left, the gradient descent vector is shown for a linear regression problem, though now the input data $\bm{x}$ has a correlation structure ($\langle \bm{x}\bm{x}^\top \rangle = \bm{\Sigma}$). Right, if the correlation matrix is integrated into the parameters, alignment with the direction of steepest descent in this landscape is returned.}
    \label{fig:NGDCorrVis}
\end{figure}

Figure~\ref{fig:NGDCorrVis} visualises this effect for a linear regression problem, equivalent to a single layer of a network.
When the data carries a correlation structure, gradient descent points off-axis and takes a detour during optimization.
Viewed in a parameterization which absorbs $\bm{\Sigma}^{1/2}$ into the parameters, the same vector once again points toward the minimum.
The right factor of Equation~\ref{eq:bg_NG_DNN} is precisely the transformation which undoes this, and from this observation arrives our main proposal: removing the input correlations at every layer of a deep neural network would improve gradient descent directions for \textbf{any} algorithm, approximate or otherwise.

There are two routes by which this may be achieved.
One may measure the correlation structure, invert it, and apply that inverse to the gradient, which is the gradient preconditioning described in Section~\ref{sec:bg_precond}.
Alternatively, one may modify the network itself, such that the data arriving at every layer is already decorrelated and ordinary gradient descent is thereby already operating in an orthonormal basis.
We pursue the second route.
In Section~\ref{sec:equiv} we show that these two routes are not merely similar but exactly equivalent, provided one additional correction is made.

\subsection{A decorrelating transform, learned online}
\label{sec:transform}

At every layer of a network we construct a linear transformation of the data, $\bar{\bm{x}}_i = \bm{M}_i\bm{x}_i$, where $\bar{\bm{x}}_i$ is intended to be a decorrelated form of the data ${\bm{x}}_i$.
A network thereby gains one additional linear transformation per layer and its computation is written
\begin{equation*}
\bm{x}_i = \phi(\bm{h}_i) =  \phi (\bm{W}_i \bar{\bm{x}}_{i-1}) = \phi (\bm{W}_i\bm{M}_{i-1}\bm{x}_{i-1}),
\end{equation*}
with all elements as described in Section~\ref{sec:bg_dnn}, and with the addition of a square decorrelating matrix at every layer.

Suppose that we wish to whiten the data, such that $\left\langle\bar {\bm x}_{i-1} \bar {\bm x}_{i-1}^\top\right\rangle = \bm I$.\footnote{In contrast, the objective of decorrelation is $\left\langle\bar {\bm x}_{i-1} \bar {\bm x}_{i-1}^\top\right\rangle = \bm D$, for an arbitrary diagonal matrix $\bm D$. Therefore, whitening is a special case of decorrelation.}
\citet{silva1991distributed} propose to minimise a whitening loss equivalent to $\mathcal L_\text{Whiten}(\bm M_i) = \frac14 \| \left\langle \bm M_i\bm x_{i}\bm x_{i}^{\top}\bm M_i^{\top}\right \rangle_{\bm x_{i}} -\bm I \|^2_F.$
Notably, this loss is itself minimised at a linear transformation, and so its minimization is subject to `being ruined by correlations' as described above.
Applying right-sided natural gradient descent to it, rather than ordinary gradient descent, we derive the update equation
\begin{equation}
    \bm M_i \leftarrow \bm M_i - \eta_M \left(\bar{\bm x}_{i} \bar{\bm x}_{i}^\top - \bm I \right) \bm M_i,
    \label{eq:M}
\end{equation}
which is similar in form to those of \citet{silva1991distributed} and \citet{Ahmad2022-hv}, and whose derivation is given in Appendix~\ref{app:derive_M}.
This update requires only the decorrelated activity and the decorrelation parameters itself, all of which are locally available.
% , and so is available wherever that activity is.

\subsection{Correcting the forward weights}
\label{sec:correction}

One issue which arises is that every update to the decorrelation matrix via Equation~\ref{eq:M} alters the function which the network computes as it alters the forward pass.
The consequence is a displacement of the network upon its task's loss landscape as we show in Appendix~\ref{app:instabilities}.
In other words, the minimisation of the whitening loss competes against the task loss.
Such an effect also disturbs any momentum terms and gradient traces upon which modern optimizers depend.
Existing methods for online decorrelation \citep{silva1991distributed, Ahmad2022-hv, Dalm2024-ly, Huang2018-zy} ignore this as a factor, and try to limit its effect by keeping $\eta_M$ small, even though one would prefer it to be as large as possible in order to track a changing correlation structure.

We propose instead to remove this conflict, by deriving the exact change in the parameters $\bm W_i$ that would ensure that any learning done for decorrelation does not affect the network's output at all, but only the basis in which that layer represents and uses its inputs.
Formally, we require that before vs after a decorrelation update, the effective map of a layer remains unchanged:
\begin{equation*}
\bm W_i^{(u)}\bm M_{i-1}^{(u)}\bm x_{i-1} = \bm W_i\bm M_{i-1}\bm x_{i-1},
\end{equation*}
where the superscript $(u)$ denotes the updated matrices.
Because the update of Equation~\ref{eq:M} is a rank-one modification when applied to a single sample, this condition may be solved by the Sherman-Morrison inverse without any matrix decomposition, giving
\begin{equation}
\bm W_i^{(u)} = \frac1\gamma\bm W_i + \frac{\eta_M \bm h_i \bar {\bm x}_{i-1}^\top}{\gamma^2 - \eta_M\gamma \bar {\bm x}_{i-1}^\top \bar {\bm x}_{i-1}},
\label{eq:W}
\end{equation}
with $\gamma = 1 + \eta_M$, and with the full derivation given in Appendix~\ref{app:decor}.
Note that this update is itself Hebbian, being an outer product of a layer's input $\bar{\bm x}_{i-1}$ with its own output $\bm h_i$ and is easily computed.

With this correction in place, the network's function is preserved exactly at every step, and the decorrelating transform is free to learn as quickly as it likes.
It is worth noting what this buys beyond stability.
Because the function computed by the network is unchanged, the transform cannot remove information which the network might otherwise have used, and the concern raised by \citet{Wadia2021-jt}, that whitening a dataset restricts the space of generalization unless carefully regularised, is mitigated.
Our networks are whitened at every layer and yet express exactly the family of functions they would have expressed without whitening.

\subsection{Three equivalent forms}
\label{sec:equiv}

We may now state the central result of this work.
Equations~\ref{eq:M} and \ref{eq:W} together describe a single mechanism which admits three implementations, and these implementations are exactly equivalent in the updates which they produce.
It is convenient to state this in terms of the \textit{effective} map of a layer, $\bm V_i = \bm W_i \bm M_{i-1}$, being the linear transformation which the layer actually applies to its input, and to write $\bm\delta_i = \partial \ell / \partial \bm h_i$ for the gradient arriving at that layer.

\begin{boxedtheorem}[Three equivalent forms]
\label{thm:equiv}
Let $\bm M_{i-1}$ be updated by Equation~\ref{eq:M} upon a single sample.
Then the three procedures below induce identical updates to the effective map $\bm V_i = \bm W_i\bm M_{i-1}$:
\begin{center}
\renewcommand{\arraystretch}{1.5}
\begin{tabular}{@{}p{0.05\linewidth}>{\raggedright\arraybackslash}p{0.38\linewidth}>{\raggedright\arraybackslash}p{0.47\linewidth}@{}}
\hline
 & \textbf{Layer input} & \textbf{Weight update} \\
\hline
(i) & $\bm h_i = \bm W_i\bm x_{i-1}$ & $\bm W_i \gets \bm W_i - \eta \, \bm\delta_i\bm x_{i-1}^\top \bm M_{i-1}^\top \bm M_{i-1}$ \\
(ii) & $\bm h_i = \bm W_i\bar{\bm x}_{i-1}$, \; $\bar{\bm x}_{i-1} = \bm M_{i-1}\bm x_{i-1}$ & $\bm W_i \gets \bm W_i - \eta \, \bm\delta_i\bar{\bm x}_{i-1}^\top$, \; then Equation~\ref{eq:W} \\
(iii) & $\bm h_i = \bm W_i\bar{\bm x}_{i-1}$, \; $\tau\dot{\bar{\bm x}}_{i-1} = \bm x_{i-1} - \bm R_{i-1}\bar{\bm x}_{i-1}$, \; $\bm R_{i-1} = \bm M_{i-1}^{-1}$ & $\bm W_i \gets \bm W_i - \eta \, \bm\delta_i\bar{\bm x}_{i-1}^\top$, \; then Equations~\ref{eq:W} and \ref{eq:R} \\
\hline
\end{tabular}
\end{center}
Whenever $\bm M_{i-1}$ additionally sits at the whitening fixed point, this common update is the right-sided natural gradient update of Equation~\ref{eq:bg_NG_DNN}.
\end{boxedtheorem}

Under (ii), gradient descent upon $\bm W_i$ gives $\delta \bm W_i = -\eta \bm\delta_i\bar{\bm x}_{i-1}^\top = -\eta \bm\delta_i\bm x_{i-1}^\top \bm M_{i-1}^\top$, so the induced change in the effective map is $\delta \bm V_i = \delta \bm W_i \bm M_{i-1} = -\eta \bm\delta_i\bm x_{i-1}^\top \bm M_{i-1}^\top\bm M_{i-1}$, which is exactly the update of (i).
Equation~\ref{eq:W} guarantees that the remaining change to $\bm V_i$, that arises from the update of $\bm M_{i-1}$ itself, is zero.
Equivalence of (ii) and (iii) follows from the fixed point of the dynamics being $\bar{\bm x}_{i-1} = \bm R_{i-1}^{-1}\bm x_{i-1} = \bm M_{i-1}\bm x_{i-1}$, with $\bm R_{i-1}$ maintaining that inverse relationship by induction.
The final statement follows from $\bm M_{i-1}^\top\bm M_{i-1} = \langle \bm x_{i-1} \bm x_{i-1}^\top\rangle^{-1}$ at the whitening fixed point.
Since the equivalence holds for the effective map of every layer, it holds for the network as a whole, and the backward pass is unaffected: propagating through $\bm V_i^\top = \bm M_{i-1}^\top\bm W_i^\top$ in (ii) and (iii) is identical to propagating through $\bm V_i^\top$ in (i).

Note, the equivalences we have outlined above strictly hold only for single-sample updates.
For a mini-batch of $B$ samples the corresponding update is of rank $\min(B, n_i)$, and its exact inverse requires the Woodbury identity together with a $B \times B$ inversion, which is no longer computable from quantities available at a single connection.
However note that the regime in which our method is exact (single-sample online updating) is also the regime in which its locality matters, as neither a nervous system nor an analogue device processes a mini-batch.

\begin{wrapfigure}{r}{0.5\linewidth}
    \vspace{-10pt}
    \centering
    \includegraphics[width=\linewidth]{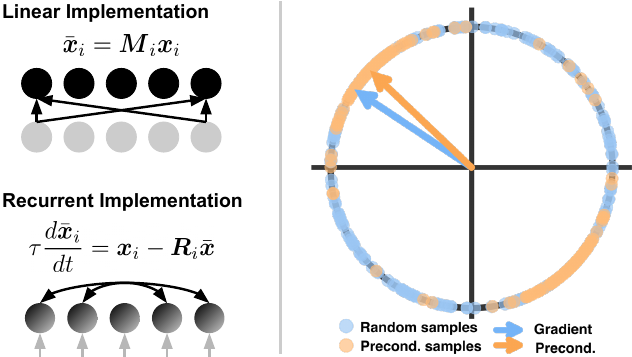}
    \caption{Left, our proposed method can be equivalently represented as either a single linear transformation or the fixed point of a recurrent set of dynamics (see the main text). Right, randomly sampled directions on a skewed quadratic loss function landscape that fall within 90 degrees of the natural/preconditioned gradient direction can be improved by our method.}
    \label{fig:decor}
    \vspace{-8pt}
\end{wrapfigure}

\paragraph{As a gradient preconditioner}
At the whitening fixed point, $\bm M_i\langle \bm x_i \bm x_i^\top\rangle \bm M_i^\top = \bm I$, and therefore $\bm M_i^\top\bm M_i = \langle \bm x_i \bm x_i^\top\rangle^{-1}$.
The transform thus carries an estimate of the inverse input correlation matrix, and one may dispense with the forward transformation altogether and write the update as
\begin{equation}
\bm{W}_{i} \gets \bm{W}_{i} - \eta_W \frac{\partial \ell}{\partial \bm h_i} \bm{x}_{i-1}^\top \bm M_{i-1}^\top\bm M_{i-1},
\label{eq:precond}
\end{equation}
which is right-sided natural gradient descent as it appears in Equation~\ref{eq:bg_NG_DNN}, with the inverse correlation supplied by $\bm M$ rather than by a matrix inversion.
Note that far from the fixed point, $\bm M_i^\top\bm M_i$ is not the inverse correlation matrix, but instead akin to a (typical) preconditioner which is learning its parameters and continuously tuning to the correlation structure of the data.

\paragraph{As a linear transform}
Alternatively, one may propagate the decorrelated activities $\bar{\bm x}_i$ through the network as in Section~\ref{sec:transform}, apply ordinary gradient descent to $\bm W_i$, and additionally apply the correction of Equation~\ref{eq:W} after every update of $\bm M_i$.
The network then performs ordinary gradient descent in a basis which is continuously tending toward being orthonormal and comes at the cost of increased computational complexity for forward passes.

\paragraph{As a system of recurrent dynamics}
Finally, the same transform may be realised as the fixed point of a set of linear recurrent dynamics,
\begin{equation}
\tau\frac{d \bar{\bm{x}}_i }{dt} = \bm{x}_i - \bm{R}_i \bar {\bm{x}}_i,
\label{eq:R_dyn}
\end{equation}
which is exactly equivalent to the linear transform above when $\bm{R}_i = {\bm{M}_i}^{-1}$.
The matrix $\bm{R}_i$ need never be inverted, but may be updated directly by the same Sherman-Morrison argument used for Equation~\ref{eq:W}, giving
\begin{equation}
\bm R_i^{(u)} = \frac1\gamma\bm R_i + \frac{\eta_M \bm x_{i} \bar{\bm x}_{i}^\top}{\gamma^2 - \eta_M\gamma\bar {\bm x}_{i}^\top \bar {\bm x}_{i}}.
\label{eq:R}
\end{equation}
Equations~\ref{eq:W} and \ref{eq:R} have the same form, and both are Hebbian.
See Figure~\ref{fig:decor}, left, for a depiction of the linear and recurrent implementations.

The equivalence of these three forms is the point of this section, and it has a consequence which we wish to state plainly.
In the recurrent form, every quantity appearing in Equations~\ref{eq:M}, \ref{eq:W} and \ref{eq:R} is available at the connection being updated: the activity entering it, and the activity leaving it.
There is no matrix inverse, no decomposition, and no gathering of a layer's statistics by machinery outside of the network.
Nor is any of it computed from the gradient, so the whole mechanism runs during the forward pass, before a backward pass exists.
Right-sided natural gradient descent is thereby available in settings where a preconditioner is not: in distributed systems in which no coordinator exists, in analogue and neuromorphic hardware in which matrix inversion is not a primitive, and in nervous systems, in which a synapse knows only what its two neurons are doing.

\paragraph{Computational cost} All three forms add in memory a square matrix of order $\mathcal{O}(n_i^2)$ per layer during training, where $n_i$ is the number of nodes in layer $i$, along with the cost of propagating inputs through it and of updating it.
For convolutional layers, note that this matrix acts upon the patch dimension rather than the spatial extent of a feature map, and so is of size $(C k^2)^2$ for $C$ channels and a $k \times k$ kernel, which is considerably smaller than a naive reading of $\mathcal{O}(n_i^2)$ would suggest.
To give an indication of wall-clock costs, we report these in Appendix~\ref{app:walltime}.
The recurrent form's efficiency is highly dependent upon the software and hardware upon which it is implemented, and one should not consider it efficient unless applied in custom neuromorphic, analogue, or other exotic hardware, nervous systems included.

%% file: sections/results.tex
\section{Results}\label{sec:results}

% Our experiments address two questions.
% First, whether removing input correlations at every layer improves credit assignment, and whether it does so most where the gradient signal is least reliable.
% Second, whether a correction which is computed locally, and during the forward pass, can compete with preconditioners which compute a comparable correction centrally and from the gradient.

% % TODO(E1): equivalence check - the three forms of Theorem 1 upon one seed, at B=1 (exact) and at
% % B=128 (measures the O(eta_M^2) residual noted in Section 3.4). Belongs here, ahead of 4.2.
% % TODO(E2): correction ablation - with and without Equation 4, momentum on.

% \subsection{Setup}
% \label{sec:setup}
We test our proposed approach by training a LeNet-like network and a five hidden-layer convolutional network upon CIFAR-10 and CIFAR-100 with backpropagation (BP), by feedback alignment (FA)~\citep{Lillicrap2016-nm}, and by uniform sign-concordance (SC)~\citep{liao2016important}.
We also test in VGG-7, VGG-9, and ResNet-18 architectures with error predictive coding (ePC)~\citep{goemaere2025epc}, a more recent local learning method which has been shown to scale to deeper networks.
Network architectures and training hyperparameters are given in Appendix~\ref{app:arch}, and pseudocode for both the preconditioning and the linear-transform implementations in Appendix~\ref{app:pseudocode}.

\begin{figure}[!ht]
    \centering
    \includegraphics[width=0.85 \linewidth]{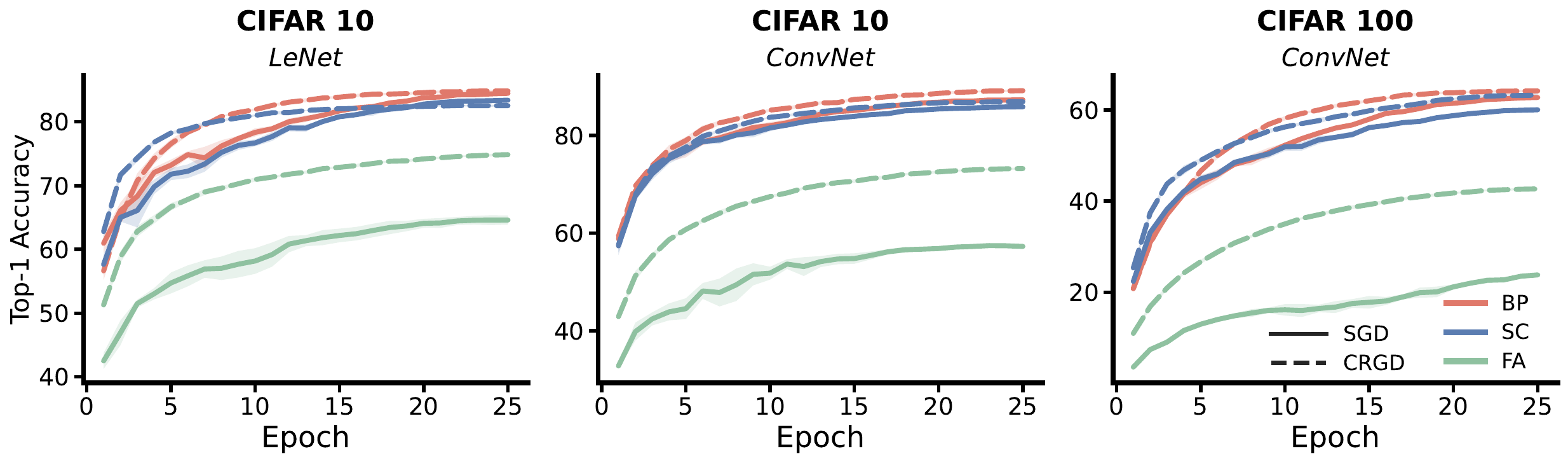}
    \caption{Computing right-sided natural gradient updates massively helps in generalisation performance of approximate backpropagation algorithms. Results are shown for backpropagation (BP), feedback alignment (FA), and  sign-concordant-fa (SC) trained upon the CIFAR10/100 in a LeNet-like and a five hidden-layer convolutional networks. Network architectures and training hyperparameters are described in detail in Appendix~\ref{app:arch}. Envelopes show the standard deviation in accuracy levels across five randomly seeded networks.}
    \label{fig:results}
\end{figure}

Figure~\ref{fig:results} shows test-set performances for all three credit assignment methods, with and without our method.
Every one of these algorithms computes its update in a basis where data correlations have made non-orthonormal, and so every one of them gains from a return to an orthonormal basis, backpropagation included.
Consistent with our proposal that the effect of right-sided gradient correction is separate to the gradient quality, the least accurate credit assignment method, feedback alignment, gains the most from our method, and the most accurate, backpropagation, gains the least.

\begin{figure}[!ht]
    \centering
    \includegraphics[width=0.85 \linewidth]{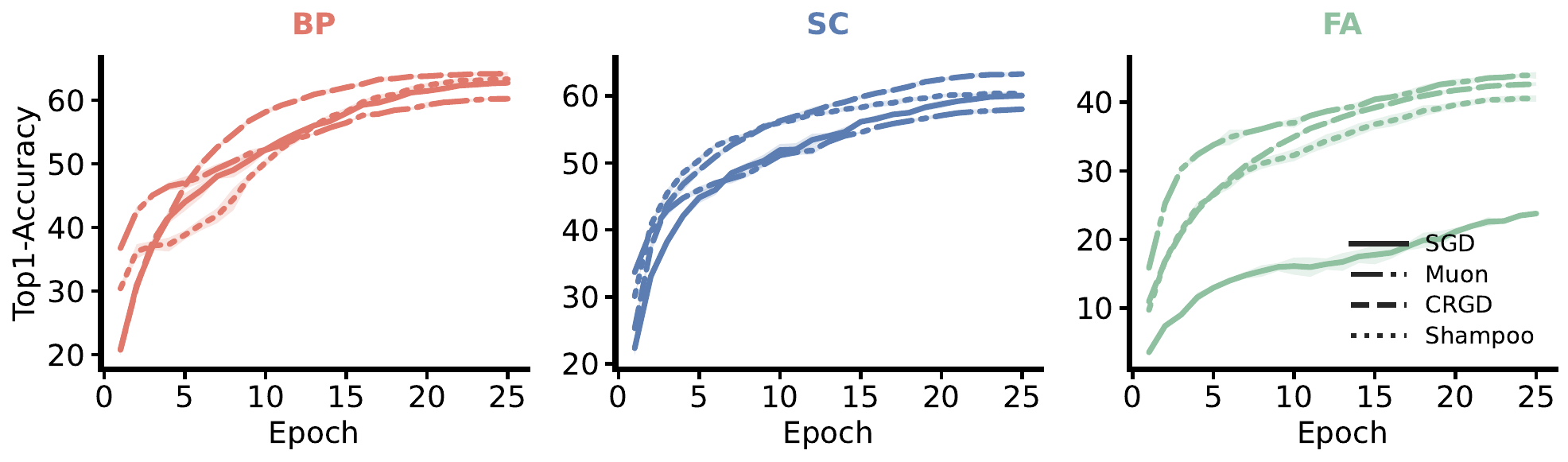}
    \caption{Our method compared against Shampoo \citep{Gupta2018-co} and Muon \citep{jordan2024muon}, for each credit assignment method, in the five hidden-layer convolutional network trained upon CIFAR-100. Envelopes show the standard deviation across five randomly seeded networks. The same comparison for every network and dataset is given in Appendix~\ref{app:optimizer_grid}. Note that our correction is computed during the forward pass and from locally available quantities alone, whereas both baselines require statistics of the gradient together with a matrix decomposition or orthogonalization carried out outside of the network.}
    \label{fig:optimizers}
    \vspace{-10pt}
\end{figure}

Section~\ref{sec:bg_precond} observed that existing preconditioners require statistics of the gradient and a centralised matrix operation, neither of which our method needs, but whose impact could be nonetheless important.
Figure~\ref{fig:optimizers} compares our method against Shampoo \citep{Gupta2018-co} and Muon \citep{jordan2024muon} in the convolutional network upon CIFAR-100, the same comparison across every network and dataset provided in Appendix~\ref{app:optimizer_grid}.

With backpropagation and with sign-concordance, our method matches or outperforms both modern optimizers. % TODO: insert numbers
We take this to be the central claim of this work: a correction which never leaves the forward pass, never inverts a matrix, and never gathers the statistics of a layer into one place is nonetheless competitive with preconditioners which do all three.
Feedback alignment is the exception, and Muon outperforms our method here.
We attribute this to the fact that feedback alignment suffers from two distinct gradient estimation difficulties: its updates suffer from a low-rank tendency in it's credit assignment subspace \citep{boeshertz2026overcoming}, and updates are additionally expressed in a non-orthonormal basis we have identified.
Our method acts strictly upon the second of these factors, while Muon can also address the gradient rank issue.
That a method targeting the gradient factor might win where the gradient factor is the dominant problem, while a method targeting the input factor wins elsewhere, precisely aligns with the separation which we create between backward-pass based and forward-pass based corrections.

\begin{table}[!ht]
\small
\centering
\caption{Top-1 test accuracy for error-based predictive coding (ePC), for ePC combined with our method, and for backpropagation, across deeper architectures. Mean and standard deviation across seeds. Bold marks the better of the two ePC variants. Backpropagation is given in grey as a reference rather than as a competing method, and is italicised where it exceeds both.}
\label{tab:epc}
\begin{tabular}{@{}llcc>{\color{gray}}c@{}}
\toprule
Model & Dataset & ePC & ePC + CRGD & \textcolor{gray}{BP} \\
\midrule
VGG-7 & CIFAR-10 & $88.84 \pm 0.31$ & $\bm{90.59 \pm 0.22}$ & \bpval{89.60 \pm 0.16} \\
      & CIFAR-100 & $58.62 \pm 0.20$ & $\bm{64.24 \pm 0.30}$ & \bplead{65.14 \pm 0.29} \\
VGG-9 & CIFAR-10 & $86.81 \pm 0.09$ & $\bm{88.85 \pm 0.27}$ & \bplead{89.76 \pm 0.20} \\
      & CIFAR-100 & $60.65 \pm 0.25$ & $\bm{63.46 \pm 0.45}$ & \bpval{61.11 \pm 0.45} \\
ResNet-18 & CIFAR-10 & $91.73 \pm 0.21$ & $\bm{92.41 \pm 0.08}$ & \bpval{91.85 \pm 0.24} \\
          & CIFAR-100 & $69.47 \pm 0.32$ & $\bm{71.26 \pm 0.23}$ & \bplead{71.46 \pm 0.32} \\
\bottomrule
\end{tabular}
\end{table}

% TODO: the numbers above are the cross-entropy columns of the NeurIPS rebuttal table. Replace
% with the newer ePC results, and decide whether the MSE columns warrant an appendix.
To establish that these effects are not confined to the small networks or simple credit assignment algorithms above, we additionally apply our method to error-based predictive coding (ePC)~\citep{goemaere2025epc} in deeper architectures.
Table~\ref{tab:epc} shows that our method improves ePC in every architecture and upon both datasets, and that the combination closes much of the remaining distance to backpropagation, exceeding it in several settings.
The margins here are of a few percent rather than the far larger ones seen above, which is consistent with predictive coding supplying a considerably better estimate of the gradient than feedback alignment does.

%% file: sections/discussion.tex
\section{Discussion}\label{sec:discussion}

Herein we have shown that the input-side factor of the natural gradient need not be computed by an optimizer at all.
It can be removed by the network itself, provided the forward weights are simultaneously corrected.
The three forms of Theorem~\ref{thm:equiv} are the consequence of this finding, and the recurrent form describes one half of natural gradients as being computable during the forward pass, from quantities present at a single connection, with no inverse and no decomposition required.
Existing preconditioners apply their correction to the gradient because that is where an optimizer has access, however one half can in fact happen, locally, in parallel, and without any gradient computation.
% TODO: cite one-sided variants (ASGO, one-sided Shampoo/Muon) and PSGD/Kron once in the bib.
% Recent one-sided methods, which precondition upon a single factor, sit closest to our own work and approach the same term from the opposite direction.
% Read in this way, the choice of exponent which distinguishes several of these methods, whether one applies $\bm\Sigma^{-1}$, $\bm\Sigma^{-1/2}$ or some intermediate power, becomes a question of \textit{where} the transformation is placed rather than of how strong it should be: a $\bm\Sigma^{-1/2}$ in the forward pass is a $\bm\Sigma^{-1}$ in the gradient, as Section~\ref{sec:equiv} makes explicit.

Our decorrelating transform is close in form to rules proposed repeatedly over three decades \citep{silva1991distributed, silva1991speeding, Ahmad2022-hv, Dalm2024-ly}, and the practice of whitening layer activations is well established \citep{Huang2018-zy, Huang2019-ms, Luo2017-ku}.
We add theoretical backing, an efficient learning rule, and a correction to the forward weights, Equation~\ref{eq:W}, which holds the network's function fixed while decorrelation is carried out and without which our recurrent method would have no preconditioner equivalent.

In computational neuroscience, lateral connections updated in proportion to the correlation of the activities they connect, so-called anti-Hebbian learning, have been proposed as a mechanism for competition and decorrelation in populations of neurons for a similar length of time \citep{Oja1989-wa, Foldiak1990-gv, Pehlevan2015-pr, Lipshutz2024-an} with experimental support in the centre-surround processing of early vision \citep{Pitkow2012-wz} and in the spatial decorrelation produced by inhibitory plasticity \citep{He2019-nd}.
% TODO: cite Duong et al. on adaptive whitening with gain-modulating interneurons.
Our results and derivation of Equation~\ref{eq:R} suggest another interpretation of this body of work: a circuit which decorrelates its own inputs is, for whatever else it may be doing, placing the synapses downstream of it into an orthonormal basis, making ordinary Hebbian-style learning at those synapses behave as natural gradient descent.
Furthermore, the update of Equation~\ref{eq:W} then adds the suggestion that such systems should also adjust a neuron's forward weights so as to hold its output fixed while the representation arriving at it is reorganised, a form of homeostatic plasticity.
That these two properties emerge from an optimization level derivation are remarkable and hint at a relationship between the optimization of credit assignment and the homeostatic and inhibitory properties of biological neural networks.

Having said all of the above, practically speaking the equivalences we derive are exact only for single-sample updates and, as discussed in Section~\ref{sec:equiv}, a mini-batch update scheme would require the computation of at least one inverse matrix.
Our method also requires the addition of a square matrix at every network layer, which for very wide and deep networks is a real cost.
Finally our evaluation is confined to convolutional and residual networks at the CIFAR task scale.
Exploring efficient scaling options for very large networks and for larger datasets, and exploring the method's applicability to other credit assignment methods, are natural next steps.

Finally, our method repairs the input factor of Equation~\ref{eq:bg_NG_DNN} and leaves the gradient factor alone, whereas methods such as Muon do the reverse, and the results of Section~\ref{sec:results} suggest the two failures to be genuinely distinct.
A method which repaired both, one within the network and one within the optimizer, seems to us the natural thing to try next though questions remain as to how it could be implemented in a biological system or neuromorphic platform.
% CUT (2026-09-25, for page count): the generality claim that other credit assignment methods
% should benefit too. Already made empirically by BP/FA/SC/ePC in the Results, and it cost ~4
% lines. Restore if space appears:
% the credit assignment methods treated here are not the only ones our method might enable, as any algorithm which produces an imperfect estimate of the gradient while leaving a layer's input statistics intact stands to benefit in the same way \citep{Nokland2016-ux, Neftci2019-tf, Dalm2023-hq, xiao2018biologically, ernoult2022towards}.

%% file: sections/appendices.tex
\clearpage
\section{Background on Gradient Descent and Natural Gradients}
\label{app:background}

\subsection{Gradient Descent}
Consider the case in which we have a dataset which provides input and output pairs $(\bm{x},\bm{y}) \in \mathcal{D}$, and we wish for some parameterised function, $\bm{z} = \bm{f}_{\boldsymbol{\theta}}(\bm{x})$, with parameters $\bm{\theta}$ to produce a mapping relating these.
This mapping would be optimal if it minimised a loss function $\mathcal{L}(\bm{\theta}) = \frac{1}{|\mathcal{D}|} \sum_{(\bm{x},\bm{y})\in\mathcal{D}} \ell (\bm{f}_{\bm{\theta}}(\bm{x}), \bm{y})$, with a sample-wise loss function $\ell$.

The problem which we wish to solve in general is to minimise our loss function and find the optimal set of parameters, effectively to find ${\operatorname{argmin}}_{\bm{\theta}} \, \mathcal{L}(\bm{\theta})$.
However, finding this minimum directly is challenging for most interesting problems.
Gradient descent proposes a first-order optimization process by which we identify an update direction (not directly the optimized value) for our parameters based upon a linearization of our loss function.
This is often formulated as taking a `small step' in the steepest descent direction of a function, in its gradient direction, such that
\begin{equation*}
\delta {\bm{\theta}}^{\text{GD}} = - \eta \nabla \mathcal{L}(\bm{\theta}),
\end{equation*}
where $\eta$ is the step size.

However, this supposition of taking a 'small step' in the gradient direction hides a specific assumption.
In fact, it is equivalent to minimising a linear approximation (1st order Taylor expansion) of our loss function with an added penalization based upon the change in our parameters.
Specifically, this is the optimum solution of
\begin{equation*}
\delta {\bm{\theta}}^{\text{GD}} = \underset{\delta \bm{\theta}}{\operatorname{argmin}} \, \mathcal{L}(\bm{\theta}) + \nabla \mathcal{L}(\bm{\theta})^\top \delta \bm{\theta} + \frac{1}{2 \eta}{\delta \bm{\theta}}^\top \delta \bm{\theta}.
\end{equation*}
where, ${\delta \bm{\theta}}^\top \delta \bm{\theta} = ||\delta \bm{\theta}||^2_2 = \sum_i \delta \bm{\theta}_i^2$.
Note, one can derive gradient descent by simply finding the minimum of this optimization function.
Thus, gradient descent assumes that it is sensible to measure and limit the distance of our parameter update in terms of the squared (Euclidean) norm of the parameter change.
Natural gradients supposes that this is not the best choice.

\subsection{Natural Gradient Descent}

\begin{figure}[!ht]
    \centering
    \includegraphics[width=0.7\linewidth]{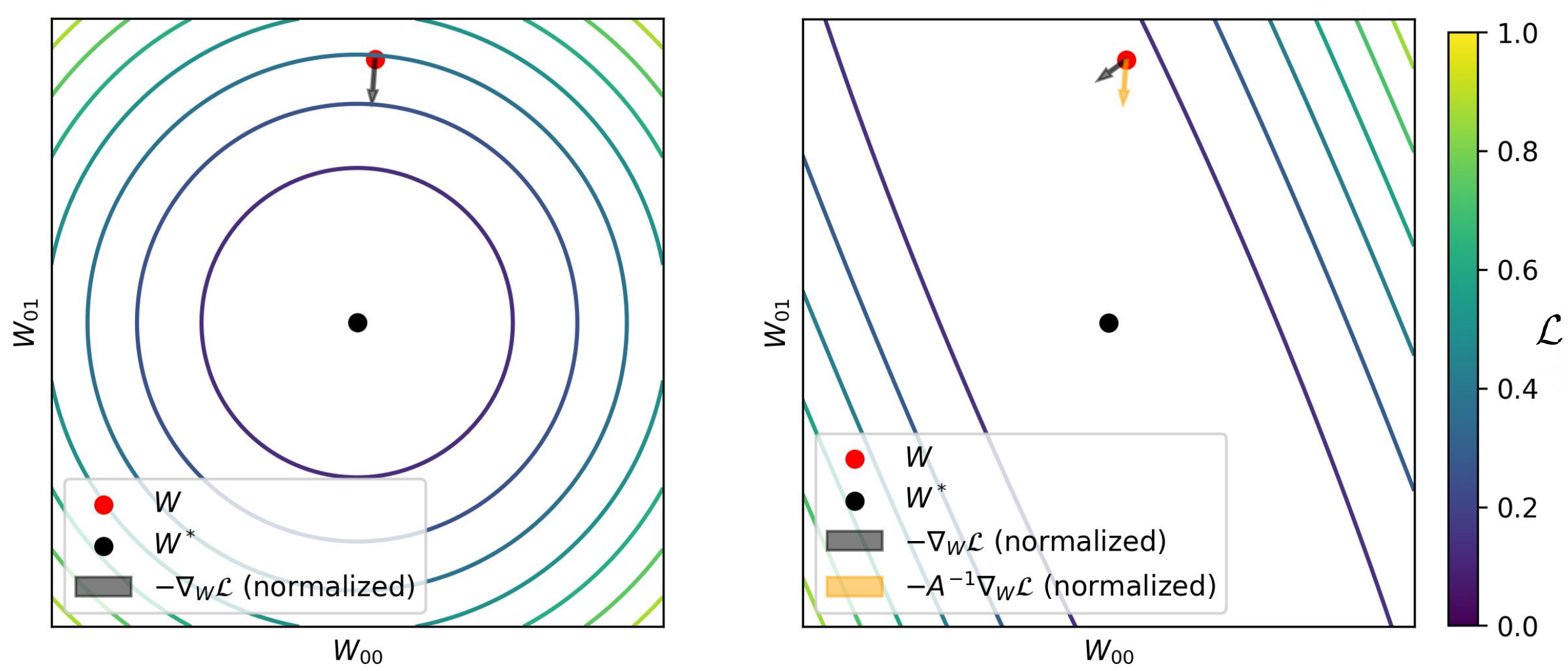}
    \caption{Left is shown a loss function landscape (in parameter space) which is a well-conditioned convex loss. In red is shown a model being optimised and the gradient descent direction (grey arrow) points toward the minimum of the function.
    Right, the loss function is modified to include a positive semi-definite scaling matrix ($\mathcal{L} = \sum_{(\bm{x},\bm{y}) \in \mathcal{X}} (\bm{W}\bm{x} - \bm{y})^\top \bm{A} (\bm{W}\bm{x} - \bm{y})$), which causes a skew and stretch of the loss landscape. Gradient descent no longer points toward the minimum but instead in the direction of steepest descent (a detour), whereas natural gradient descent (orange arrow) points directly once more at the loss functions minimum.}
    \label{fig:NGDVis}
\end{figure}

\citet{Amari1998-wh}, proposed the concept of natural gradients, and put forth the Riemannian metric as the sensible alternative to a regular Euclidean measurement of distance.
Specifically, rather than define squared distance with the Euclidean metric ($|\delta \bm{\theta}|^{\text{Euclidean}^2} = \sum_i \delta \bm{\theta}_i^2$), you can instead use a general metric instead, 
\begin{equation*}
|\delta \bm{\theta}|^{\text{Riemannian}} = \sum_i \sum_j g_{ij} \delta \bm{\theta}_i \delta \bm{\theta}_j = \delta {\bm{\theta}}^\top \bm{G}(\bm{\theta}) \delta \bm{\theta}
\end{equation*}
where $\bm{G}$ is the Riemannian metric matrix.
Note, that the values of the matrix, $\bm{G}$, are a function of $\bm{\theta}$ and are thus not static.
We can now make use of our Riemannian metric in place of the previous Euclidean metric, and by finding the minimum of this optimization, we can arrive at a neat formulation of natural gradient descent,
\begin{equation*}
{\delta \bm{\theta}}^{\text{NGD}} = - \eta \bm{G}^{-1} \nabla \mathcal{L}(\bm{\theta}).
\end{equation*}

Natural gradient descent aims to ensure that it is not the steepest direction of descent which is taken, but instead the direction which undoes any skew in the loss function to arrive at a more direct descent toward (local) minima.
This difference is illustrated in Figure~\ref{fig:NGDVis}.

\clearpage
\section{Loss distance: The appropriate Riemann metric for natural gradients}\label{app:lossdist}
In the main text, we propose that one may find the appropriate form of the Riemann metric for gradient descent by supposing that the metric measured should tend to the total summed sample-wise loss distance
$$
\delta \theta^{\top} G \delta \theta \rightarrow \frac{1}{|\mathcal{D}|} \sum_{(x_0,y) \in \mathcal{D}} (\ell(f(\theta + \delta \theta, x_0), y) - \ell(f(\theta, x_0), y))^2.
$$
By expanding the loss function with it's Taylor series,
\begin{align}
(\ell(f(\theta + \delta \theta, x_0), y) - \ell(f(\theta, x_0), y))^2 &= (\ell + \nabla_{\theta} \ell \cdot {\delta \theta} + {\delta \theta}^{\top} \nabla^2_{\theta} \ell {\delta \theta} + ... - \; \ell)^2 \\
&= ({\delta \theta}^\top \nabla_{\theta} \ell  + {\delta \theta}^\top \nabla^2_{\theta} \ell {\delta \theta} + ...)^2 \\
&= {\delta \theta}^{\top} \nabla_{\theta} \ell \nabla_{\theta} \ell^{\top}  {\delta \theta} + 2(\nabla_{\theta} \ell {\delta \theta} \times {\delta \theta}^{\top} \nabla^2_{\theta} \ell {\delta \theta}) + ... \\
\end{align}
As the change in parameters tend to small values (or equivalently as $\lim_{\eta \rightarrow 0}$), this term can be approximated by its first term.
This first term, when used as the measure, is thus a simple equivalence to the Riemann metric
$$
{\delta \theta}^{\top} G \delta \theta = {\delta \theta}^{\top} \left \langle \nabla_{\theta} \ell \nabla_{\theta} \ell^{\top} \right \rangle_{x_0 \in \mathcal{D}}  {\delta \theta}.
$$

\clearpage
\section{Derivation for the Riemann metric for a deep neural network}\label{app:dnn_natgrads}
Defining our model as
$$
x_i = \phi(h_i) =  \phi (W_i x_{i-1})
$$
for $i \in [1 ... L]$, where $L$ is the number of layers in our network, $\phi$ is a non-linear activation function (which we shall assume is a fixed continuous transfer function for all layers), $W_i$ is a matrix of parameters for layer $i$, $x_0$ is our input data, and $x_L$ is our model output.
We ignore biases for now as a simplification of our derivation.

First let us begin by collecting all parameter matrices into a single vector of parameters, such that $\theta = \text{Vec}(W_1, W_2, ... W_L)^\top$ where $\text{Vec}(\cdot)$ indicates the flattening of a tensor into a vector.
Next we can define, the derivative of our loss with respect to the parameters,
$$
\nabla_\theta \mathcal{L} = \left( \frac{\partial \mathcal{L}}{\partial \text{Vec}(W_1, W_2, .. W_L)} \right) = \left \langle
\text{Vec} \left (\frac{\partial \ell}{\partial h_1} x_0^\top, \frac{\partial \ell}{\partial h_2} x_1^\top , ...\frac{\partial \ell}{\partial h_L}  x_{L-1}^\top \right) \right \rangle_{x_0 \in \mathcal{D}}
$$
where we have replaced the derivative terms with the layer computation as can be calculated by backpropagation.
We can now compute more straightforwardly the outer product of this gradient vector with itself, such that
\begin{align*}
G(\theta) &= \left \langle \nabla_{\theta} \ell \nabla_{\theta} \ell^{\top} \right \rangle_{x_0}\\
 &=  \left \langle \text{Vec} \left (\frac{\partial \ell}{\partial h_1} x_0^\top, \frac{\partial \ell}{\partial h_2} x_1^\top , ...\frac{\partial \ell}{\partial h_L}  x_{L-1}^\top \right)
 \text{Vec} \left (\frac{\partial \ell}{\partial h_1} x_0^\top, \frac{\partial \ell}{\partial h_2} x_1^\top , ...\frac{\partial \ell}{\partial h_L}  x_{L-1}^\top \right) ^\top  \right \rangle_{x_0}
\end{align*}
This is now a matrix of shape $G(\theta) \in \mathbb{R}^{M \times M}$ assuming $\theta \in \mathbb{R}^{M}$.

Rather than making explicit how to calculate the entire matrix, is is more fruitful to break down this computation into the separate blocks of this term, such that the $(i,j)$-block of our matrix is computed
\begin{equation*}
G_{ij} = \left\langle  \text{Vec} \left( \frac{\partial \ell}{\partial h_j} x_{j-1}^\top \right ) \text{Vec} \left ( \frac{\partial \ell}{\partial h_i} x_{i-1}^\top \right )^\top \right\rangle_{x_0} \\
= \left\langle  \left( x_{j-1} \otimes \frac{\partial \ell}{\partial h_j} \right ) \left ( x_{i-1} \otimes \frac{\partial \ell}{\partial h_i} \right )^\top \right\rangle_{x_0} \\
\end{equation*}
where $\otimes$ represents the Kronecker (Zehfuss) product. The mixed-product property of the Kronecker product allows us to also re-formulate this as
\begin{equation*}
G_{ij} = \left \langle \left (x_{j-1} x_{i-1}^\top \right ) \otimes \left ( \frac{\partial \ell}{\partial h_j} \frac{\partial \ell}{\partial h_i}^\top  \right) \right \rangle_{x_0}
= \left \langle x_{j-1} x_{i-1}^\top \right \rangle_{x_0} \otimes \left \langle \frac{\partial \ell}{\partial h_j} \frac{\partial \ell}{\partial h_i}^\top \right \rangle_y.
\end{equation*}
Note that we here made a separation of the expectation values based upon the fact that the loss value depends entirely on label or target output, $y$, and the network activities depend entirely on the value of the inputs, $x_0$. We assume, as \citet{Bernacchia2018-zj} do, that these can therefore be separately averaged.

There are now two possible ways to proceed, each requiring an alternative assumption. First, one may suppose that we decide to optimise only a single weight matrix of the network at a time, which would allow us to ignore all but the diagonal blocks of our matrix, $G$.
\citet{Bernacchia2018-zj} alternatively arrived at a sole consideration of the (inverse) diagonal blocks of $G$ by assuming a linear network (i.e. that $\phi(x) = x$), and showing that in such a case, the matrix $G$ is singular and that it's pseudo-inverse can be taken using only the diagonal blocks.

Regardless, if we take a single block at a time, and suppose that we are only updating single weight matrices, we can now formulate the natural gradients learning rule as

\begin{align*}
{\delta \theta}_{W_i} &= - {\eta}\langle \nabla_{\theta_{W_i}} \ell \, \nabla_{\theta_{W_i}} \ell^\top \rangle_{x_0}^{-1} \nabla_{\theta_{W_i}} \mathcal{L} \\
&= - {\eta} \left( \left \langle x_{i-1} x_{i-1}^\top \right \rangle_{x_0} \otimes \left \langle \frac{\partial \ell}{\partial h_i} \frac{\partial \ell}{\partial h_i}^\top \right \rangle_{y} \right )^{-1} \nabla_{\theta_{W_i}} \mathcal{L} \\
&= - {\eta} \left( \left \langle x_{i-1} x_{i-1}^\top \right \rangle^{-1}_{x_0} \otimes \left \langle \frac{\partial \ell}{\partial h_i} \frac{\partial \ell}{\partial h_i}^\top \right \rangle^{-1}_{y} \right ) \text{Vec} \left \langle \frac{\partial \ell}{\partial h_i} x_{i-1}^\top \right \rangle_{x_0} \\
&= - {\eta} \text{Vec} \left( \left \langle \frac{\partial \ell}{\partial h_i} \frac{\partial \ell}{\partial h_i}^\top \right \rangle^{-1}_{y} \left \langle \frac{\partial \ell}{\partial h_i} x_{i-1}^\top \right \rangle_{x_0} \left \langle x_{i-1} x_{i-1}^\top \right \rangle^{-1}_{x_0} \right ) \\
\end{align*}

Thus, we arrive at an expression which allows us to better interpret the impact of computing the natural gradient. We can appreciate that the natural gradient formulation thus achieves two things:
- The left matrix multiplication removes any skew and mis-scaling of the loss function with respect to the hidden activities, dealing with the problem that we classically associate with a skewed loss landscape
- The right matrix multiplication removes the impact of any \textit{correlation structure in our input data}. Correlation structure in our input-data is equivalent to our parameters living in a non-orthonormal basis set and thus affects the speed and accuracy of training!

\clearpage
\section{Minimizing the whitening loss via right-sided gradient descent}
\label{app:derive_M}
We wish to minimize the loss 
\begin{equation*}
    \mathcal L_\text{Whiten}(\bm M_i) = \frac14 \| \left\langle \bm M_i\bm x_{i}\bm x_{i}^{\top}\bm M_i^{\top}\right \rangle_{\bm x_{i}} -\bm I \|^2_F,
\end{equation*}
without being ruined by data correlations.
Therefore, our update equation will be of the form 
\begin{equation*}
    \bm M_i \leftarrow \bm M_i - \eta_M \nabla \mathcal L_\text{Whiten} (\bm M_i) \langle \bm x_{i} \bm x_{i} \rangle^{-1}_{\bm x_{i}}
\end{equation*}

First, we take the derivative of the loss with respect to $\bm M_i$:
\begin{align*}
    \nabla \mathcal L_\text{Whiten}(\bm M_i) &= \left(\langle \bm M_i\bm x_i \bm x_i^\top \bm M_i^\top\rangle_{\bm x_i} - \bm I\right) \langle \bm M_i \bm x_i \bm x_i^\top \rangle_{\bm x_i} \\
    &= \left( \bm M_i\langle\bm x_i \bm x_i^\top\rangle_{\bm x_i} \bm M_i^\top - \bm I\right) \bm M_i \langle \bm x_i \bm x_i^\top\rangle_{\bm x_i} \\
\end{align*}

Filling the gradient back in gives us the update equation
\begin{align*}
\bm M_i \leftarrow &\bm M_i - \eta_M \nabla \mathcal L_\text{Whiten}(\bm M_i)\langle \bm x_{i} \bm x_{i}^\top \rangle^{-1}_{\bm x_{i}}\\
=& \bm M_i - \eta_M \left( \bm M_i\langle\bm x_i \bm x_i^\top\rangle_{\bm x_i} \bm M_i^\top - \bm I\right) \bm M_i \langle \bm x_i \bm x_i^\top\rangle_{\bm x_i} \langle \bm x_{i} \bm x_{i}^\top \rangle^{-1}_{\bm x_{i}}\\ 
=& \bm M_i - \eta_M \left( \bm M_i\langle\bm x_i \bm x_i^\top\rangle_{\bm x_i} \bm M_i^\top - \bm I\right) \bm M_i\\ 
\end{align*}

This update equation requires us to aggregate the data correlations before updating. In practice, we use stochastic gradient descent to relax the equation to arrive at our stochastic update
\begin{equation*}
\bm M_i \leftarrow \bm M_i - \eta_M \left( \bm M_i \bm x_i \bm x_i^\top \bm M_i^\top - \bm I\right) \bm M_i.
\end{equation*}

\section{Our decorrelation mechanism as a recurrent system of dynamics}\label{app:decor}

We desire $\bm W_i^{(u)}\bm M_{i-1}^{(u)}\bm x_{i-1} = \bm W_i\bm M_{i-1}\bm x_{i-1} $, where the superscript $(u)$ denotes the updated matrices. In the following, we will drop the subscripts as it should be obvious from the context.
We begin the derivation using Equation~\ref{eq:M}
\begin{align*}
    \bm W^{(u)}\bm M^{(u)}\bm x &= \bm W\bm M\bm x\\
    \bm W^{(u)}\bm M^{(u)} &= \bm W\bm M\\
    \bm W^{(u)}(\bm M - \eta_M (\bar{\bm x} \bar{\bm x}^\top - \bm I ) \bm M) &= \bm W\bm M\\
    \bm W^{(u)} &= \bm W\bm M\left(\bm M - \eta_M (\bar{\bm x} \bar{\bm x}^\top - \bm I ) \bm M\right)^{-1} \\
    \bm W^{(u)} &= \bm W\bm M\left((1+\eta_M)\bm I - \eta_M\bar{\bm x} \bar{\bm x}^\top) \bm M\right)^{-1} \\
    \bm W^{(u)} &= \bm W\left((1+\eta_M)\bm I - \eta_M\bar{\bm x} \bar{\bm x}^\top\right)^{-1} \\
\end{align*}

In this form, we can apply the Sherman-Morrison inverse.
\begin{align*}
\left((1+\eta_M)\bm I - \eta_M\bar{\bm x} \bar{\bm x}^\top\right)^{-1} 
= &\frac1{(1+\eta_M)}\bm I + \eta_M\frac{\frac{1}{(1+\eta_M)^2}\bar{\bm x} \bar{\bm x}^\top} {1 - \frac{\eta_M}{1+\eta_M}\bar{\bm x}^\top \bar{\bm x}}\\
= &\frac1{(1+\eta_M)}\bm I + \eta_M\frac{\bar{\bm x} \bar{\bm x}^\top} {(1+\eta_M)^2 - \eta_M(1+\eta_M)\bar{\bm x}^\top \bar{\bm x}}\\
\end{align*}

We can now define $\gamma = 1 + \eta_M$ to slightly simplify this equation and plug it back in.
\begin{align*}
    \bm W^{(u)} &= \bm W\left((1+\eta_M)\bm I - \eta_M\bar{\bm x} \bar{\bm x}^\top\right)^{-1} \\
    &= \bm W\left(\frac1\gamma\bm I + \eta_M\frac{\bar{\bm x} \bar{\bm x}^\top} {\gamma^2 - \eta_M\gamma\bar{\bm x}^\top \bar{\bm x}}\right)\\
    &= \frac1\gamma\bm W + \eta_M\frac{\bm h \bar{\bm x}^\top} {\gamma^2 - \eta_M\gamma\bar{\bm x}^\top \bar{\bm x}}
\end{align*}

Note that in the last step, we have used that $\bm h_i = \bm W_i\bm x_{i-1}$. This finishes the derivation for Equation~\ref{eq:W}.

To derive Equation~\ref{eq:R}, we can use a similar approach.
We iteratively want to update $\bm R_i$ to be equivalent to $\bm M_i$ at every point during the training process. 
Therefore, we initialize the two matrices as the identity matrix, so that the equivalence holds at the first time-step. 
Then, we make an inductive argument to derive the update equation to ensure that this equivalence holds after updating $\bm M_i$.

\begin{align*}
    \bm R^{(u)}_i &= \bm (M^{(u)}_i)^{-1} \\
    &= \left(\bm M - \eta_M (\bar{\bm x} \bar{\bm x}^\top - \bm I ) \bm M\right)^{-1}\\
    &= \left((1+\eta_M)\bm I - \eta_M\bar{\bm x} \bar{\bm x}^\top) \bm M\right)^{-1}\\
    &= \bm R\left((1+\eta_M)\bm I - \eta_M\bar{\bm x} \bar{\bm x}^\top\right)^{-1}
\end{align*}

We can reuse the derivation of the inverse using Sherman-Morrison to derive the update equation.
\begin{align*}
\bm R^{(u)}_i &= \bm R\left((1+\eta_M)\bm I - \eta_M\bar{\bm x} \bar{\bm x}^\top\right)^{-1}\\
&= \bm R\left(\frac1\gamma\bm I + \eta_M\frac{\bar{\bm x} \bar{\bm x}^\top} {\gamma^2 - \eta_M\gamma\bar{\bm x}^\top \bar{\bm x}}\right)\\
&= \frac1\gamma\bm R + \eta_M\frac{\bm x \bar{\bm x}^\top} {\gamma^2 - \eta_M\gamma\bar{\bm x}^\top \bar{\bm x}}
\end{align*}

\clearpage
\section{Problems of iterative decorrelation}
\label{app:instabilities}
Here, we want to demonstrate that updating a decorrelation matrix, without accounting for the change in the network leads to instabilities during convergence.

We consider a linear neural network parameterized by $\bm W_1$ that maps from multiple inputs to a single output, and decorrelation matrix $\bm M_0$. Further, we consider a simple mean-squared error loss function 
$\mathcal L(\bm W_1, \bm M_0, \mathcal D) =  \frac12\left\langle (\bm W_1\bm M_0\bm x_0 - \bm y)^2\right\rangle_{\bm x_0, \bm y \in D}$. 
We are interested in the loss difference between this network, and the network upon updating the decorrelation matrix denoted by $\bm M_0^{(u)} = \bm M_0 + \delta \bm M_0$.
\begin{align*}
\epsilon &= L(\bm W_1, \bm M_0, \mathcal D) - L(\bm W_1, \bm M^{(u)}_0, \mathcal D)\\
&= \frac12\left\langle (\bm W_1\bm M_0\bm x_0 - \bm y)^2 - (\bm W_1(\bm M_0+\delta\bm M_0)\bm x_0 - \bm y)^2\right\rangle_{\bm x_0, \bm y \in D}.\\
&= \frac12\left\langle (\bm W_1\bm M_0\bm x_0 - \bm y)^2 - ((\bm W_1 \bm M_0\bm x_0 - \bm y) + (\bm W_1 \delta\bm M_0\bm x_0))^2\right\rangle_{\bm x_0, \bm y \in D}.\\
    &= \frac12\left\langle 2(\bm W_1\bm M_0\bm x_0 - \bm y)(\bm W_1\delta\bm M_0\bm x_0) -  (\bm W_1\delta\bm M_0\bm x_0)^2\right\rangle_{\bm x_0, \bm y \in D}\\
    &= \left\langle (\bm W_1\bm M_0\bm x_0 - \bm y)(\bm W_1\delta\bm M_0\bm x_0) -  \frac12(\bm W_1\delta\bm M_0\bm x_0)^2\right\rangle_{\bm x_0, \bm y \in D}\\
    &= \left\langle (\bm W_1\bm M_0\bm x_0 - \bm y)(\bm W_1\delta\bm M_0\bm x_0) -  \frac12(\bm W_1\delta\bm M_0\bm x_0)^2\right\rangle_{\bm x_0, \bm y \in D}\\
    &= \bm W_1\bm M_0 \bm \Sigma \delta \bm M_0^\top \bm W_1^\top - \bm c\delta \bm M_0^\top \bm W_1^\top -  \frac12\bm W_1\delta\bm M_0 \bm \Sigma \delta \bm M_0^\top \bm W_1^\top,
\end{align*}

where $\bm \Sigma = \langle \bm x_0\bm x_0^\top\rangle_{\bm x_0\in \mathcal D}$, and $\bm c = \langle \bm x_0\bm y\rangle_{\bm x_0, \bm y\in \mathcal D}$.

What we can see is that as long as our decorrelation updates are not the zero matrix, a loss is induced in our network propagated by the downstream matrices and our matrix change. 
Both the sign and magnitude of the loss difference are under-determined, but depend on the weights, the decorrelation matrix, and the inputs and outputs of the network. 
In other words, we have no control over where the model ends up after a decorrelation update. We can only tone down the influence through the decorrelation learning rate, although we, at the same time, wish for it to be as high as possible to drive down the decorrelation loss.

As a consequence, this form of decorrelation without adjusting the weights leads to a strong dependence on careful tuning of the learning rate. 
Furthermore, we expect that very deep networks may be very unstable to train. Not only due to the highly fluctuating inputs to every intermediate layer, but also due to the momentum terms commonly used in optimization that are becoming inaccurate after every decorrelation update.

\clearpage
\clearpage
\section{Architectures and Hyperparameters}\label{app:arch}

The networks trained for the experiments of Figure~\ref{fig:results} are of two types: a LeNet-like convolutional architecture and a deeper five hidden-layer convolutional architecture.
Both are given in Table~\ref{tab:architectures}.
Training and testing use the CIFAR-10 and CIFAR-100 datasets~\citep{Krizhevsky09learningmultiple}, each composed of a training set of 50,000 images, split into either 10 or 100 classes, and a test set of 10,000 images.
We apply the standard data augmentations for the training set: random crops to $32 \times 32$ pixels with padding $=4$, random horizontal flips, and normalization.

\begin{table}[!ht]
\caption{The neural network architectures for the experiments in Figure~\ref{fig:results}. For convolutional and pooling layers, shapes are organised as `Kernel Width $\times$ Kernel Height $\times$ Output Channels (Stride, Padding)'. Each parameterized layer is followed by an exponential linear unit (ELU) activation function~\citep{clevert2015fast}.}
\label{tab:architectures}
\centering
\begin{scriptsize}
\begin{sc}
\renewcommand{\arraystretch}{1.5} % Increases space between rows
\begin{tabular}{l|l|l}
{Network} &  {Layer Types} & {Layer size} \\
 \midrule
\multirow{7}{*}{LeNet Architecture} &  Input & $32\times32\times3$  \\
  & Conv & $5 \times 5 \times 32, (1, 2)$ \\
  & MaxPool & $3 \times 3, (2, 1)$\\
  & Conv & $5 \times 5 \times 32, (1, 2)$\\
  & MaxPool & $3 \times 3, (2, 1)$\\
  & Linear & 512 \\
  & Linear & 10 \\
\hline
\multirow{12}{*}{ConvNet Architecture (CIFAR10/100)} &  Input  & 32$\times$32$\times$3\\
  & Conv & $3\times3\times128, (1, 1)$ \\
  & MaxPool & $2\times2 (2, 0)$  \\
  & Conv & $3\times3\times128, (1, 1)$ \\
  & MaxPool & $2\times2 (2, 0)$  \\
  & Conv & $3\times3\times256, (1, 1)$ \\
  & MaxPool & $2\times2 (2, 0)$  \\
  & Conv & $3\times3\times256, (1, 1)$ \\
  & MaxPool & $2\times2 (2, 0)$  \\
  & Conv & $3\times3\times512, (1, 1)$ \\
  & MaxPool & $2\times2 (2, 0)$  \\
  & FC & 10 (or 100 for CIFAR 100)\\
  \hline
 \end{tabular}
\end{sc}
\end{scriptsize}
\end{table}

\begin{table}[!ht]
\caption{Ranges searched for each optimizer. Every hyperparameter was drawn from a log-uniform distribution over the interval given, with 100 trials for each combination of network, credit assignment method and optimizer. A dash marks a hyperparameter which does not apply.}
\label{tab:sweepranges}
\centering
\begin{scriptsize}
\begin{sc}
\renewcommand{\arraystretch}{1.2}
\begin{tabular}{@{}lccc@{}}
\toprule
{Optimizer} & {Learning rate} & {Weight decay} & {Additional} \\
\midrule
SGD & $[10^{-5}, 5\!\times\!10^{-1}]$ & $[10^{-5}, 10^{-3}]$ & -- \\
CRGD (ours) & $[10^{-5}, 5\!\times\!10^{-1}]$ & $[10^{-5}, 10^{-3}]$ & $\eta_M \in [10^{-7}, 10^{-4}]$ \\
Shampoo & $[10^{-5}, 5\!\times\!10^{-1}]$ & $[10^{-5}, 10^{-3}]$ & -- \\
Muon & $[10^{-4}, 5\!\times\!10^{-1}]$ & $[10^{-6}, 10^{-2}]$ & AdamW LR $\in [10^{-4}, 10^{-1}]$ \\
\bottomrule
\end{tabular}
\end{sc}
\end{scriptsize}
\end{table}

\begin{table}[!ht]
\caption{Hyperparameters selected for the simulations of Figures~\ref{fig:results} and \ref{fig:optimizers}, for every combination of network, credit assignment method and optimizer. A dash marks a hyperparameter which does not apply to that optimizer: the decorrelation learning rate $\eta_M$ exists only for our method, and the AdamW learning rate only for Muon, which applies AdamW to those parameters it does not orthogonalize. Values are given to three significant figures.}
\label{tab:hyperparams}
\centering
\begin{scriptsize}
\begin{sc}
\renewcommand{\arraystretch}{1.15}
\begin{tabular}{@{}lllcccc@{}}
\toprule
{Dataset, Network} & {Method} & {Optimizer} & {Learning rate} & {Weight decay} & {Decor. LR} & {AdamW LR} \\
\midrule
\multirow{12}{*}{CIFAR-10, LeNet} & \multirow{4}{*}{BP} & SGD & $2.12\!\times\!10^{-2}$ & $4.90\!\times\!10^{-4}$ & -- & -- \\
 &  & CRGD (ours) & $3.86\!\times\!10^{-2}$ & $2.14\!\times\!10^{-4}$ & $5.51\!\times\!10^{-6}$ & -- \\
 &  & Shampoo & $3.89\!\times\!10^{-2}$ & $7.34\!\times\!10^{-4}$ & -- & -- \\
 &  & Muon & $1.14\!\times\!10^{-2}$ & $1.97\!\times\!10^{-3}$ & -- & $2.12\!\times\!10^{-3}$ \\
\cmidrule(l){2-7}
 & \multirow{4}{*}{SC} & SGD & $2.77\!\times\!10^{-2}$ & $5.91\!\times\!10^{-4}$ & -- & -- \\
 &  & CRGD (ours) & $1.80\!\times\!10^{-2}$ & $5.74\!\times\!10^{-5}$ & $4.05\!\times\!10^{-5}$ & -- \\
 &  & Shampoo & $3.51\!\times\!10^{-2}$ & $8.07\!\times\!10^{-4}$ & -- & -- \\
 &  & Muon & $1.24\!\times\!10^{-2}$ & $2.92\!\times\!10^{-5}$ & -- & $2.53\!\times\!10^{-3}$ \\
\cmidrule(l){2-7}
 & \multirow{4}{*}{FA} & SGD & $5.80\!\times\!10^{-3}$ & $5.11\!\times\!10^{-4}$ & -- & -- \\
 &  & CRGD (ours) & $4.13\!\times\!10^{-2}$ & $2.88\!\times\!10^{-4}$ & $9.73\!\times\!10^{-5}$ & -- \\
 &  & Shampoo & $7.65\!\times\!10^{-3}$ & $6.34\!\times\!10^{-4}$ & -- & -- \\
 &  & Muon & $3.07\!\times\!10^{-3}$ & $6.26\!\times\!10^{-6}$ & -- & $2.76\!\times\!10^{-4}$ \\
\midrule
\multirow{12}{*}{CIFAR-10, ConvNet} & \multirow{4}{*}{BP} & SGD & $3.16\!\times\!10^{-2}$ & $1.92\!\times\!10^{-5}$ & -- & -- \\
 &  & CRGD (ours) & $4.52\!\times\!10^{-2}$ & $5.40\!\times\!10^{-4}$ & $1.69\!\times\!10^{-5}$ & -- \\
 &  & Shampoo & $1.96\!\times\!10^{-2}$ & $6.43\!\times\!10^{-4}$ & -- & -- \\
 &  & Muon & $7.65\!\times\!10^{-3}$ & $6.25\!\times\!10^{-4}$ & -- & $2.39\!\times\!10^{-3}$ \\
\cmidrule(l){2-7}
 & \multirow{4}{*}{SC} & SGD & $1.98\!\times\!10^{-2}$ & $8.37\!\times\!10^{-4}$ & -- & -- \\
 &  & CRGD (ours) & $3.82\!\times\!10^{-2}$ & $4.10\!\times\!10^{-4}$ & $9.29\!\times\!10^{-7}$ & -- \\
 &  & Shampoo & $4.22\!\times\!10^{-2}$ & $3.84\!\times\!10^{-4}$ & -- & -- \\
 &  & Muon & $5.89\!\times\!10^{-3}$ & $4.05\!\times\!10^{-4}$ & -- & $8.73\!\times\!10^{-4}$ \\
\cmidrule(l){2-7}
 & \multirow{4}{*}{FA} & SGD & $1.05\!\times\!10^{-2}$ & $8.17\!\times\!10^{-4}$ & -- & -- \\
 &  & CRGD (ours) & $3.09\!\times\!10^{-2}$ & $1.49\!\times\!10^{-5}$ & $8.49\!\times\!10^{-5}$ & -- \\
 &  & Shampoo & $1.49\!\times\!10^{-2}$ & $6.85\!\times\!10^{-5}$ & -- & -- \\
 &  & Muon & $1.91\!\times\!10^{-3}$ & $1.31\!\times\!10^{-3}$ & -- & $1.43\!\times\!10^{-4}$ \\
\midrule
\multirow{12}{*}{CIFAR-100, ConvNet} & \multirow{4}{*}{BP} & SGD & $3.94\!\times\!10^{-2}$ & $8.95\!\times\!10^{-4}$ & -- & -- \\
 &  & CRGD (ours) & $4.13\!\times\!10^{-2}$ & $9.54\!\times\!10^{-4}$ & $1.61\!\times\!10^{-6}$ & -- \\
 &  & Shampoo & $5.24\!\times\!10^{-2}$ & $7.77\!\times\!10^{-4}$ & -- & -- \\
 &  & Muon & $6.60\!\times\!10^{-3}$ & $1.85\!\times\!10^{-4}$ & -- & $7.42\!\times\!10^{-3}$ \\
\cmidrule(l){2-7}
 & \multirow{4}{*}{SC} & SGD & $4.32\!\times\!10^{-2}$ & $9.58\!\times\!10^{-4}$ & -- & -- \\
 &  & CRGD (ours) & $5.63\!\times\!10^{-2}$ & $9.97\!\times\!10^{-4}$ & $7.71\!\times\!10^{-5}$ & -- \\
 &  & Shampoo & $5.14\!\times\!10^{-2}$ & $4.63\!\times\!10^{-4}$ & -- & -- \\
 &  & Muon & $5.25\!\times\!10^{-3}$ & $8.79\!\times\!10^{-3}$ & -- & $5.13\!\times\!10^{-3}$ \\
\cmidrule(l){2-7}
 & \multirow{4}{*}{FA} & SGD & $7.55\!\times\!10^{-3}$ & $1.20\!\times\!10^{-5}$ & -- & -- \\
 &  & CRGD (ours) & $3.11\!\times\!10^{-2}$ & $1.87\!\times\!10^{-5}$ & $7.52\!\times\!10^{-5}$ & -- \\
 &  & Shampoo & $2.00\!\times\!10^{-2}$ & $1.64\!\times\!10^{-5}$ & -- & -- \\
 &  & Muon & $2.70\!\times\!10^{-3}$ & $6.69\!\times\!10^{-3}$ & -- & $4.16\!\times\!10^{-4}$ \\
\bottomrule
\end{tabular}
\end{sc}
\end{scriptsize}
\end{table}

\begin{table}[!ht]
\caption{Hyperparameters for the error-based predictive coding experiments of Table~\ref{tab:epc}, for both the cross-entropy and mean-squared-error objectives. The learning rate and weight decay are those reported by \citet{goemaere2025epc} and are shared by both the ePC baseline and ePC + CRGD; the decorrelation learning rate $\eta_M$ applies to ePC + CRGD alone. Values are given to three significant figures.}
\label{tab:epchyperparams}
\centering
\begin{scriptsize}
\begin{sc}
\renewcommand{\arraystretch}{1.15}
\begin{tabular}{@{}lllccc@{}}
\toprule
{Model} & {Dataset} & {Loss} & {Learning rate} & {Weight decay} & {Decor. LR $\eta_M$} \\
\midrule
\multirow{4}{*}{VGG-7} & CIFAR-10 & Cross-entropy & $1.56\!\times\!10^{-3}$ & $5.46\!\times\!10^{-4}$ & $2.50\!\times\!10^{-4}$ \\
 & CIFAR-10 & MSE & $4.26\!\times\!10^{-4}$ & $2.16\!\times\!10^{-6}$ & $2.50\!\times\!10^{-4}$ \\
 & CIFAR-100 & Cross-entropy & $3.13\!\times\!10^{-4}$ & $7.99\!\times\!10^{-4}$ & $2.50\!\times\!10^{-4}$ \\
 & CIFAR-100 & MSE & $4.02\!\times\!10^{-4}$ & $1.47\!\times\!10^{-5}$ & $1.00\!\times\!10^{-4}$ \\
\midrule
\multirow{4}{*}{VGG-9} & CIFAR-10 & Cross-entropy & $5.36\!\times\!10^{-4}$ & $6.88\!\times\!10^{-4}$ & $7.50\!\times\!10^{-5}$ \\
 & CIFAR-10 & MSE & $6.61\!\times\!10^{-4}$ & $4.01\!\times\!10^{-5}$ & $2.50\!\times\!10^{-4}$ \\
 & CIFAR-100 & Cross-entropy & $3.23\!\times\!10^{-4}$ & $4.03\!\times\!10^{-4}$ & $1.00\!\times\!10^{-4}$ \\
 & CIFAR-100 & MSE & $2.01\!\times\!10^{-4}$ & $2.62\!\times\!10^{-6}$ & $2.50\!\times\!10^{-4}$ \\
\midrule
\multirow{4}{*}{ResNet-18} & CIFAR-10 & Cross-entropy & $3.39\!\times\!10^{-3}$ & $1.51\!\times\!10^{-6}$ & $1.00\!\times\!10^{-4}$ \\
 & CIFAR-10 & MSE & $7.65\!\times\!10^{-4}$ & $1.82\!\times\!10^{-4}$ & $2.50\!\times\!10^{-4}$ \\
 & CIFAR-100 & Cross-entropy & $3.03\!\times\!10^{-3}$ & $1.20\!\times\!10^{-5}$ & $7.50\!\times\!10^{-5}$ \\
 & CIFAR-100 & MSE & $3.67\!\times\!10^{-4}$ & $7.30\!\times\!10^{-4}$ & $7.50\!\times\!10^{-5}$ \\
\bottomrule
\end{tabular}
\end{sc}
\end{scriptsize}
\end{table}

The training hyper-parameters which were not searched over were held fixed across all simulations.
We used a mini-batch size of 128 and trained for 25 epochs, scheduling the learning rate with a cosine scheduler which decays it from its selected value to one hundredth of that value over the course of training.
All model weights were initialized using the Kaiming-Uniform scheme, and decorrelation weights as the identity.
The forward weights were optimized by SGD with a momentum of 0.9, except where an alternative optimizer is named in Table~\ref{tab:hyperparams}, and the cross-entropy loss was used throughout except where stated otherwise.

For every combination of network, credit assignment method and optimizer we ran a separate random search of 100 trials, drawing each hyperparameter independently from a log-uniform distribution over the ranges of Table~\ref{tab:sweepranges} and selecting the best configuration upon a held-out validation set.
% TODO: Section~\ref{sec:results} still calls this a "parameter grid search", which the ranges
% here contradict. Two words, but it is in the main text.
All searches were run from a single seed.
The selected values are given in Table~\ref{tab:hyperparams}, and those for the error-based predictive coding experiments in Table~\ref{tab:epchyperparams}.

Those hyperparameters which were not searched over were held fixed as follows.
Forward parameters carried a momentum of 0.9, whereas the decorrelation matrices were updated with neither momentum nor weight decay.
Muon was applied to the hidden weight matrices alone, and not to biases nor to the input and output layers of the network, with the remaining parameters updated by AdamW at the searched learning rate, with $\beta = (0.9, 0.95)$ and a weight decay of $0.01$.
Shampoo was run with $\beta = (0, 0.999)$, $\epsilon = 10^{-12}$, a maximum preconditioner dimension of $8192$, SGD grafting, and classical momentum of $0.9$ without dampening or Nesterov acceleration.
Its preconditioner was recomputed once every 100 steps, which is the amortisation discussed in the Introduction: between those refreshes the transform which Shampoo applies is that of a layer's gradient statistics as they stood up to 100 updates previously.
Final results were thereafter produced upon the full training and test sets, with five randomly seeded network models for each curve.
For BP, SC and FA we adapted the implementation of \citet{xiao2018biologically}.
The predictive coding models were taken directly from the implementation of \citet{goemaere2025epc}, whose choices we follow throughout.
For these experiments we adopted the learning rates and weight decays reported by \citet{goemaere2025epc} rather than searching over them ourselves, and applied them unchanged to both the ePC baseline and to ePC + CRGD.
The two therefore differ in the presence of our decorrelating transform and in nothing else, and the only quantity we selected was the decorrelation learning rate $\eta_M$, by a grid search over $[2.5\!\times\!10^{-4}, 10^{-4}, 7.5\!\times\!10^{-5}, 5\!\times\!10^{-5}, 2.5\!\times\!10^{-5}, 10^{-5}]$.
% TODO: confirm whether the backpropagation column of Table~\ref{tab:epc} was run here or is
% likewise quoted from Goemaere et al.; if quoted, its caption should say so.
We would note that these are custom architectures rather than the VGG and ResNet models as they are conventionally constructed, and we retain their names only in order that our results be comparable with those reported alongside that implementation.
All training was conducted upon a single A100 GPU, with per-epoch costs for every optimizer reported in Appendix~\ref{app:walltime}.
% NOTE: the previous draft gave total runtimes of "around 5 minutes without decorrelation and 27
% minutes with it". Those divide by the per-epoch figures of app:walltime to roughly 75-100
% epochs, so they belong to the older 100-epoch runs and cannot be right for 25. Removed rather
% than recomputed; reinstate with measured totals if wanted.
We note that our code is not optimized for runtime efficiency, and we would expect that the runtime of our decorrelation mechanism can be improved substantially.

\clearpage
\section{Comparison with preconditioned optimizers, across all networks and datasets}
\label{app:optimizer_grid}

\begin{figure}[!ht]
    \centering
    \includegraphics[width=0.95 \linewidth]{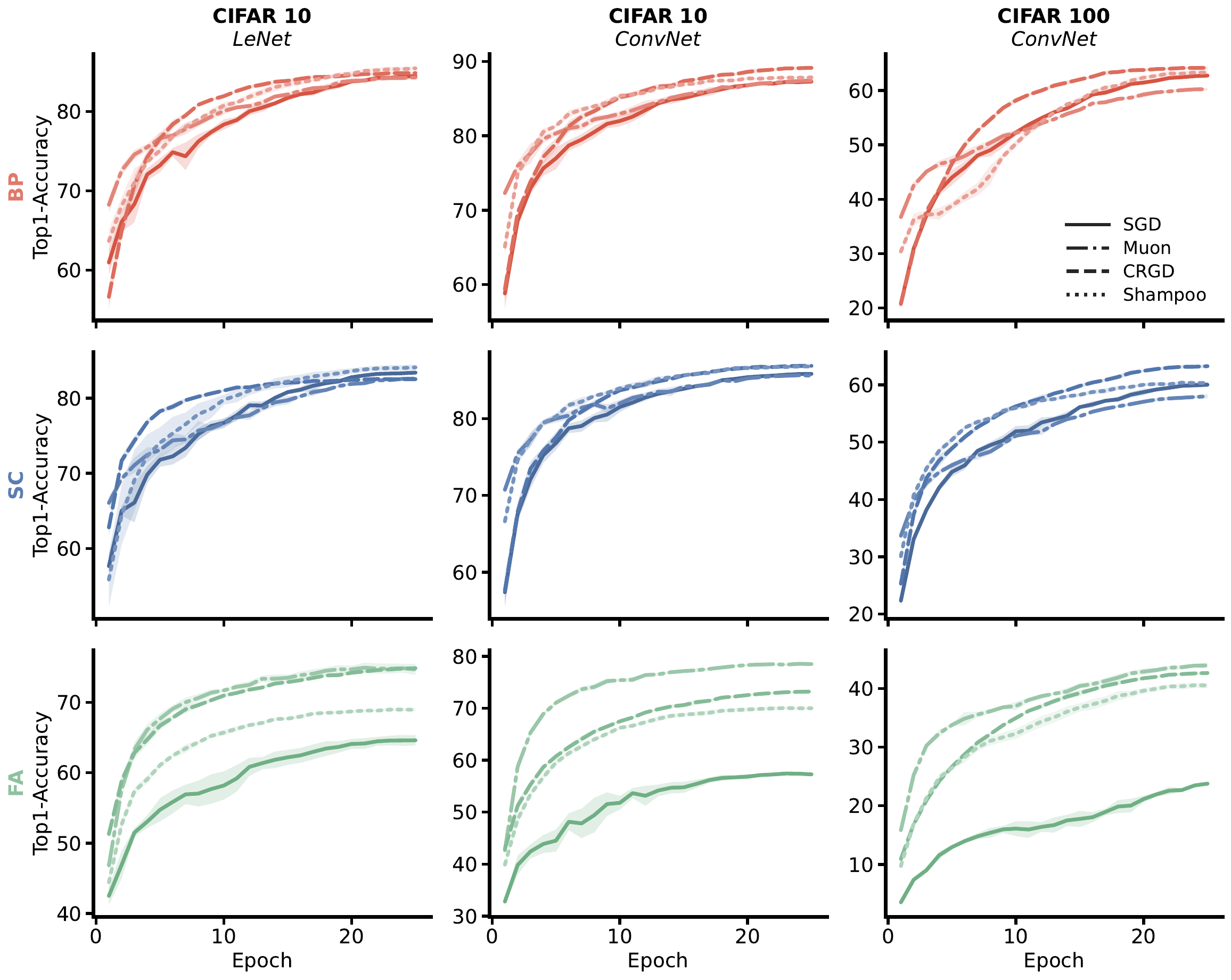}
    \caption{Our method compared against Shampoo \citep{Gupta2018-co} and Muon \citep{jordan2024muon}, for each credit assignment method (rows) and each network and dataset (columns). Envelopes show the standard deviation across five randomly seeded networks. Note that our correction is computed during the forward pass and from locally available quantities alone, whereas both baselines require statistics of the gradient together with a matrix decomposition or orthogonalization carried out outside of the network.}
    \label{fig:optimizers_full}
\end{figure}

Figure~\ref{fig:optimizers} of Section~\ref{sec:results} shows the comparison against Shampoo and Muon for the convolutional network upon CIFAR-100 alone.
Figure~\ref{fig:optimizers_full} gives the same comparison for every network and dataset which we train.
The pattern described in the main text holds throughout: our method matches or outperforms both baselines under backpropagation and under sign-concordance, whereas Muon outperforms it under feedback alignment, where the low-rank pathology of the fixed feedback matrices, rather than the input basis, is the dominant problem.

\clearpage
\section{Wall-clock cost}
\label{app:walltime}

Table~\ref{tab:walltime} reports the measured time per training epoch for each optimizer, upon CIFAR-10.
% TODO: state which of the two networks these timings are from (LeNet-like or the five
% hidden-layer convolutional network), and add the FA row if it is available.
Our method costs approximately four times ordinary gradient descent per epoch, which is appreciably less than Shampoo and appreciably more than Muon.

Two remarks are due upon these numbers.
The first is that our implementation applies the decorrelating transform as a dense matrix multiplication upon a GPU, which is the least favourable setting for it: none of the distributability which the locality of the update affords is expressed in such a measurement, and the comparison shown here is therefore one of a local method implemented centrally against methods which are centralised by construction.
The second is that the cost is borne during training alone.
The transform of a trained network may be absorbed into the forward weights of the layer which follows it, leaving only the effective map $\bm V_i = \bm W_i \bm M_{i-1}$ of Section~\ref{sec:equiv}, and inference therefore proceeds at exactly the cost of the original architecture.

\begin{table}[!ht]
\centering
\caption{Wall-clock time per training epoch, in seconds, upon CIFAR-10 (Conv). Mean and standard deviation across five randomly seeded networks. Ordinary gradient descent is given in grey as a reference point rather than as a competing method.}
\label{tab:walltime}
\begin{tabular}{@{}l>{\color{gray}}ccc c@{}}
\toprule
Credit assignment & \textcolor{gray}{SGD} & CRGD (ours) & Shampoo & Muon \\
\midrule
Backpropagation   & \bpval{4.04 \pm 0.17} & $17.60 \pm 0.13$ & $23.21 \pm 0.10$ & $5.79 \pm 0.16$ \\
Sign-concordance  & \bpval{4.16 \pm 0.24} & $17.96 \pm 0.13$ & $23.42 \pm 0.08$ & $5.72 \pm 0.43$ \\
\bottomrule
\end{tabular}
\end{table}

\clearpage
\section{Training pseudocode}\label{app:pseudocode}
\begin{algorithm}[!ht]
\caption{Right-sided natural gradient descent through preconditioning}\label{alg:precond}
\begin{algorithmic}
\renewcommand{\algorithmicrequire}{\textbf{Input:}}
 \renewcommand{\algorithmicensure}{\textbf{Output:}}
\Require {Input data} $\bm{x}_0$, {Target Output} $y$, {Forward Weights} $\bm{W}_l$, {Decorrelation Weights} $\bm{M}_l$, {Total Number of Layers} $L$, {Forward Learning Rate} $\eta_W$, {Decorrelation Learning Rate} $\eta_M$
% \Ensure {{Network Output} $\bm{x}_L$}

\State \texttt{}
\State \texttt{Compute forward pass}
\For{$i \in [0, 1, .. L-1]$}
            \State $\bar{\bm{x}}_i = \bm{M}_i \bm{x}_i$
            \Comment{Decorrelate data at layer $i$}
            \State $\bm{x}_{i+1} = \phi( \bm{W}_{i+1} {\bm{x}}_i )$
            \Comment{Pass forward to the next layer}
\EndFor

\State \texttt{}
\State \texttt{Compute credit assignment method (Example BP)}
\For{$i \in [L, L-1, ... 1]$}
    \If {$i == L$}
        \State $\bm{\delta}_L = \phi'(\bm{x}_{L}) (\bm{x}_L - \bm{y})$
        \Comment{Compute gradient at output layer (regression or CCE)}
    \ElsIf{}
        \State $\bm{\delta}_i = \phi'(\bm{x}_{i})\bm{W}_{i+1}^\top \bm{\delta}_{i+1}$
        \Comment{Backpropagate (or replace $\bm{W}_{i+1}$ for FA)}
    \EndIf
\EndFor

\State \texttt{}
\State \texttt{Update Forward Parameters}
\For{$i \in [0, 1, .. L-1]$}
            \State $\bm{W}_{i+1} \gets \bm{W}_{i+1} - \eta_W \bm{\delta}_{i+1} \bm{x}_{i}^\top \bm M_i^\top\bm M_i$
            \Comment{Precondition, then update by an optimizer}
\EndFor

\State \texttt{}
\State \texttt{Update Decorrelation Parameters}
\For{$i \in [0, 1, .. L-1]$}
            \State $\bm{C} = \bar{\bm{x}}_i \bar{\bm{x}}_i^\top$
            \Comment{Compute remaining correlation}
            % \State $\gamma = 1 + \eta_M$
            % \State $\bm{g} = \gamma^2 - \eta_M\gamma(\bar {\bm x}_i^\top \bar{\bm x}_i)$
            % \Comment{Compute rescaling}
            \State $\bm{M}_i \gets \bm{M}_i - \eta_M (\bm{C} - \bm I)\bm M_i$
            \Comment{Update decorrelation matrix}
\EndFor
\end{algorithmic}
\end{algorithm}

\begin{algorithm}
\caption{Right-sided natural gradient descent through a Linear Decorrelation Transformation}
\label{alg:linear}
\begin{algorithmic}
\renewcommand{\algorithmicrequire}{\textbf{Input:}}
 \renewcommand{\algorithmicensure}{\textbf{Output:}}
\Require {Input data} $\bm{x}_0$, {Target Output} $y$, {Forward Weights} $\bm{W}_l$, {Decorrelation Weights} $\bm{M}_l$, {Total Number of Layers} $L$, {Forward Learning Rate} $\eta_W$, {Decorrelation Learning Rate} $\eta_M$
% \Ensure {{Network Output} $\bm{x}_L$}

\State \texttt{}
\State \texttt{Compute forward pass}
\For{$i \in [0, 1, .. L-1]$}
            \State $\bar{\bm{x}}_i = \bm{M}_i \bm{x}_i$
            \Comment{Decorrelate data at layer $i$}
            \State $\bm{h}_{i+1} = \bm{W}_{i+1} \bar{\bm{x}}_i$
            \State $\bm{x}_{i+1} = \phi( \bm h_{i+1} )$
            \Comment{Pass forward to the next layer}
\EndFor

\State \texttt{}
\State \texttt{Compute credit assignment method (Example BP)}
\For{$i \in [L, L-1, ... 1]$}
    \If {$i == L$}
        \State $\bm{\delta}_L = \phi'(\bm{x}_{L}) (\bm{x}_L - \bm{y})$
        \Comment{Compute gradient at output layer (regression or CCE)}
    \ElsIf{}
        \State $\bm{\delta}_i = \phi'(\bm{x}_{i})\bm{M}_{i}^\top\bm{W}_{i+1}^\top \bm{\delta}_{i+1}$
        \Comment{Backpropagate (or replace $\bm{W}_{i+1}$ for FA)}
    \EndIf
\EndFor

\State \texttt{}
\State \texttt{Update Decorrelation Parameters}
\For{$i \in [0, 1, .. L-1]$}
            \State $\bm{C} = \bar{\bm{x}}_i \bar{\bm{x}}_i^\top$
            \Comment{Compute remaining correlation}
            \State $\gamma = 1 + \eta_M$
            \State $\bm{g} = \gamma^2 - \eta_M\gamma(\bar {\bm x}_i^\top \bar{\bm x}_i)$
            \Comment{Compute rescaling}
            \State $\bm{M}_i \gets \bm{M}_i - \eta_M (\bm{C} - \bm I)\bm M_i$
            \State $\bm W_{i+1} \gets \frac1\gamma\bm W_{i+1} + \eta_M {\bm h_{i+1}\bar{\bm x}_i^\top}/\bm g $
            \Comment{Update decorrelation matrix}
\EndFor

\State \texttt{}
\State \texttt{Update Forward Parameters}
\For{$i \in [0, 1, .. L-1]$}
            \State $\bm{W}_{i+1} \gets \bm{W}_{i+1} - \eta_W \bm{\delta}_{i+1} \bm{x}_{i}^\top$
            \Comment{Precondition, then update by an optimizer}
\EndFor

\end{algorithmic}
\end{algorithm}

Algorithm~\ref{alg:precond} describes pseudocode for the preconditioning variant for updating a network which is carried out in this work, whereas Algorithm~\ref{alg:linear} shows the pseudocode for the linear transformation variant with equivalent dynamics.
This is shown assuming the presentation of a single sample with its corresponding network update, though is in-practice used with a mini-batch of size 128 by default.
The algorithm is shown for the case of backpropagation, though it is compatible with alternative credit assignment methods such as feedback alignment or difference target propagation.